\documentclass[lettersize,journal]{IEEEtran}
\usepackage{amsmath,amsfonts}
\usepackage{algorithmic}
\usepackage{algorithm}
\usepackage{array}
\usepackage[caption=false,font=normalsize,labelfont=sf,textfont=sf]{subfig}
\usepackage{textcomp}
\usepackage{stfloats}
\usepackage{url}
\usepackage{verbatim}
\usepackage{graphicx}
\usepackage{siunitx}
\usepackage{cite}
\usepackage{amsmath}
\usepackage{makecell}
\usepackage{multirow}
\usepackage{booktabs}
\usepackage{bbm}
\usepackage{amssymb}
\usepackage{xcolor}
\definecolor{Green}{HTML}{00A64F}

\usepackage{dcolumn}

\newcommand{\ra}[1]{\renewcommand{\arraystretch}{#1}}

\begin{document}

\title{From Age Estimation to Age-Invariant Face Recognition: Generalized Age Feature Extraction Using Order-Enhanced Contrastive Learning}

\author{Haoyi~Wang,~\IEEEmembership{Member,~IEEE,}
        Victor~Sanchez,~\IEEEmembership{Member,~IEEE,}
        Chang-Tsun~Li,~\IEEEmembership{Senior~Member,~IEEE}
        Nathan~Clarke,~\IEEEmembership{Senior~Member,~IEEE}
\thanks{H. Wang and N. Clarke are with the School of Engineering, Computing and Mathematics, University of Plymouth, Plymouth, PL4 8AA, UK (e-mail: haoyi.wang@plymouth.ac.uk, N.Clarke@plymouth.ac.uk).}
\thanks{V. Sanchez is with the Department
of Computer Science, University of Warwick, Coventry, CV4 7AL, UK (e-mail: v.f.sanchez-silva@warwick.ac.uk).}
\thanks{C-T. Li is with the School of Information Technology, Deakin University, Geelong VIC 3216, Australia (e-mail: changtsun.li@deakin.edu.au).}}

\markboth{Journal of \LaTeX\ Class Files,~Vol.~14, No.~8, August~2021}%
{Shell \MakeLowercase{\textit{et al.}}: A Sample Article Using IEEEtran.cls for IEEE Journals}


\maketitle

\begin{abstract}
Generalized age feature extraction is crucial for age-related facial analysis tasks, such as age estimation and age-invariant face recognition (AIFR). Despite the recent successes of models in homogeneous-dataset experiments, their performance drops significantly in cross-dataset evaluations. Most of these models fail to extract generalized age features as they only attempt to map extracted features with training age labels directly without explicitly modeling the natural ordinal progression of aging. In this paper, we propose Order-Enhanced Contrastive Learning (OrdCon), a novel contrastive learning framework designed explicitly for ordinal attributes like age. Specifically, to extract generalized features, OrdCon aligns the direction vector of two features with either the natural aging direction or its reverse to model the ordinal process of aging. To further enhance generalizability, OrdCon leverages a novel soft proxy matching loss as a second contrastive objective, ensuring that features are positioned around the center of each age cluster with minimal intra-class variance and proportionally away from other clusters. By modeling the ageing process, the framework can enhance generalizability by improving the alignment of samples from the same class and reducing the divergence of direction vectors. We demonstrate that our proposed method achieves comparable results to state-of-the-art methods on various benchmark datasets in homogeneous-dataset evaluations for both age estimation and AIFR. In cross-dataset experiments, OrdCon outperforms other methods by reducing the mean absolute error by approximately 1.38 on average for the age estimation task and boosts the average accuracy for AIFR by 1.87\%.
\end{abstract} 


\section{Introduction}

\IEEEPARstart{A}{ge}-related facial analysis is a key area of research in biometrics and computer vision, focusing on two primary tasks: age estimation and age-invariant face recognition (AIFR). Age estimation involves predicting an individual’s chronological age from facial features, with applications in fields such as targeted marketing, surveillance, and user profiling, where age information is critical for contextually relevant decisions \cite{guo2009human, choi2011age, hu2016facial, xie2019chronological, fu2010age, wang2020using}. In contrast, AIFR aims to identify individuals across different ages, maintaining robust recognition despite changes in facial appearance over time \cite{li2011discriminative, chen2015face, shakeel2019deep, zhao2020towards}. This capability is crucial for security, forensics, and biometric authentication, where accurate tracking of individuals despite age-related changes is essential \cite{park2010age, sawant2019age}. 

Given the difficulties of handling variations in facial images captured in real scenarios, especially the diverse aging patterns across individuals, both age estimation and AIFR rely on robust and generalized age feature extraction \cite{sawant2019age}. In particular, age estimation focuses on leveraging age features directly, while AIFR requires minimizing the impact of age variations to prioritize identity features.

Despite the importance of generalized age features, current age feature extraction methods mainly rely on conventional supervised learning, where models are designed to capture facial features associated with aging by directly mapping them to training labels \cite{xia2020multi}, learning particular features tied to the age distributions and demographic characteristics in the training data \cite{rothe2018deep}. A common approach involves using deep neural networks to automatically learn age features, such as skin texture, wrinkles, and changes in facial structure \cite{wang2022improving, korban2023taa}. While these methods have demonstrated success in controlled datasets and environments, they fall short of generalizing to diverse scenarios with unfamiliar variability in aging patterns across individuals \cite{akbari2020distribution, akbari2021does}, yielding poor alignment of samples from the same class on unseen data \cite{huang2023towards}.

Additionally, models trained with conventional supervised learning techniques often struggle to capture the inherent continuity and progression of age-related changes \cite{ramanathan2009age}. Most supervised approaches treat each sample independently, disregarding the ordinal nature of the aging process \cite{rothe2018deep, wang2022improving}. This inability to capture smooth transitions in facial features over time can result in inconsistent or unrealistic predictions that do not reflect the continuous nature of aging \cite{othmani2020age}.

\begin{figure}[!t]
\centering
\includegraphics[width=3.2in]{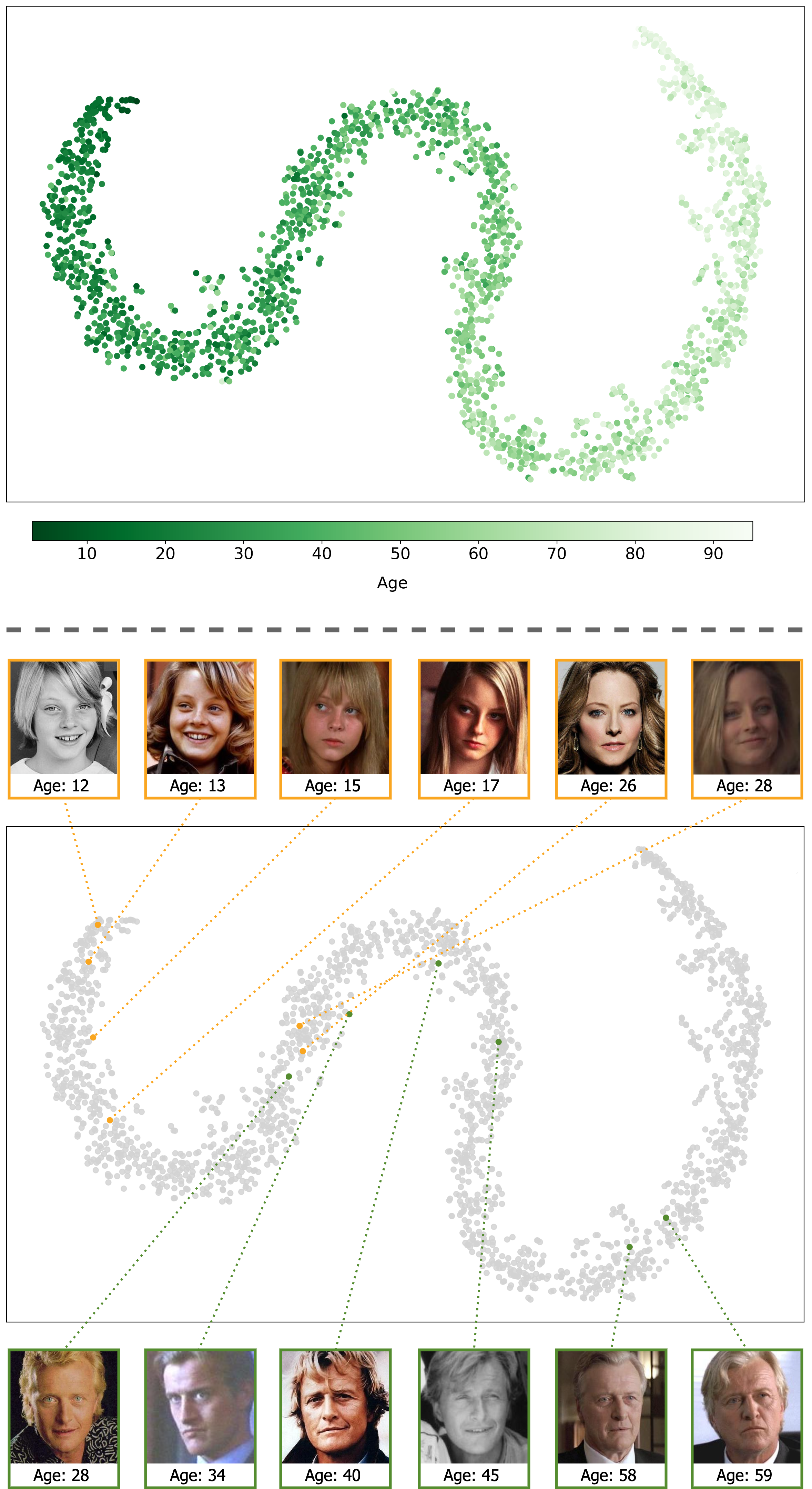}
\caption{Visual representation of the age features learned using the proposed method. Top: Feature space aligned with the direction of age progression. Bottom: Features from two individuals at different ages, following the learned age progression. Best viewed in color.}
\label{fig:demo}
\end{figure}

To address the aforementioned limitations, we propose a new contrastive learning framework called \textbf{Ord}er-Enhanced \textbf{Con}trastive Learning (OrdCon). OrdCon enhances the generalizability of age feature extraction in two ways. First, it leverages order learning \cite{lim2019order} to model how features evolve with age progression explicitly. Unlike other contrastive learning frameworks that contrast either image pairs \cite{chen2020simple, khosla2020supervised} or absolute semantic differences between samples \cite{zha2024rank}, OrdCon contrasts the direction vector between features in the feature space by aligning the direction vectors from younger to older faces with the natural direction of age progression. This approach enables the model to capture continuous and gradual transitions in facial features, effectively addressing the limitation of conventional methods in modeling progression.

Secondly, to complement the directional alignment of features along the aging progression and further boost the generalizability, OrdCon leverages a novel soft proxy matching loss as the second contractive objective to reduce intra-class variance and create a structured feature space. In this approach, proxies are learnable entities that represent the centers of specific age clusters. This loss contrasts extracted features relative to these proxies, effectively minimizing intra-class variance by ensuring that features of the same age are closely clustered. Moreover, it assigns weights to negative pairs based on their absolute age differences, thereby pushing negative pairs with greater age differences farther apart than those with smaller differences to model the relationships between features and proxies proportionally. As illustrated in Fig. \ref{fig:demo}, the top plot shows learned age features mapped in alignment with age progression, while the bottom plot demonstrates how features from the same individual at different ages are positioned in the direction of age progression.

The resulting framework for extracting generalized age features is inherently versatile. It can be directly applied to age estimation by learning an age feature space that aligns facial features along the direction of aging. For AIFR, the method builds upon our preliminary work \cite{wang2024cross}, Cross-Age Contrastive Learning (CACon), by simultaneously learning both age and identity features. To ensure that identity features are robust to age variations, the gradient for age information is gradually reversed using a gradient reversal layer (GRL) \cite{ganin2015unsupervised}. The key distinction between OrdCon and CACon lies in the explicit modeling of age progression, which is critical for improving the generalizability of both age estimation and AIFR tasks.

Since this paper focuses on applications, the theoretical analysis of the generalizability of contrastive learning for regression problems is out of scope and could be written as a separate paper. To demonstrate the generalizability of our framework, in addition to its performance in cross-dataset experiments, we adapt the matrix alignment of positive samples proposed in \cite{huang2023towards} to measure the similarity of augmented samples of the same class after contrastive learning. In addition, inspired by another matrix, the divergence of class centers proposed in \cite{huang2023towards}, we propose a new matrix called the divergence of direction vectors to better align with the objective of regression problems.

Our contributions can be summarized as follows: 
\begin{itemize} 
\item We introduce a new contrastive learning framework called \textbf{Ord}er-Enhanced \textbf{Con}trastive Learning (OrdCon) for robust and generalized age feature extraction. Unlike conventional contrastive learning methods that contrast image pairs, OrdCon contrasts direction vectors between feature pairs along the aging direction, as well as age features with age-specific proxies. 
\item We propose a soft proxy matching loss that assigns weights to negative pairs based on absolute age differences, thereby allowing for reduced intra-class variance and proportionally separated clusters for improved generalizability. 
\item To the best of our knowledge, this paper is the first to study and quantify the generalizability for models dealing with ordinal data.
\item We demonstrate that the proposed method is versatile, as it can be applied to both age estimation and AIFR tasks. 
\item We provide empirical evidence that OrdCon achieves state-of-the-art or comparable performance in homogeneous-dataset experiments, and superior performance in cross-dataset settings. 
\end{itemize}

The rest of this paper is organized as follows: Section II reviews related work on contrastive learning and order learning techniques, followed by a discussion of related works on age estimation and AIFR. In Section III, we provide a detailed description of OrdCon, including problem formulation, the definition of contrastive objectives, and its adaptation for age estimation and AIFR. Section IV presents the experimental settings and compares the performance of OrdCon to state-of-the-art methods in both homogeneous and cross-dataset experiments. Section V discusses the applicability and limitations of OrdCon, and Section VI concludes our work.

\section{Related Work}

Given that the primary focus of OrdCon is feature extraction, this section reviews key works in two core feature extraction techniques, contrastive learning and order learning, upon which our work is built. Then, we review domain-specific methods on age estimation and AIFR.

\subsection{Contrastive Learning frameworks}

Contrastive learning has emerged as a dominant and highly effective paradigm for representation learning. Among early contributions, SimCLR \cite{chen2020simple} demonstrated its effectiveness by learning a feature space where representations of positive pairs (i.e., semantically similar samples) are pulled closer together. In contrast, representations of negative pairs (i.e., dissimilar samples) are pushed far apart. SimCLR requires a large batch size for training. Its samples are generated using a composite data augmentation workflow where positive pairs originate from the same input image, while negative pairs are derived from different samples, even if they share the same class label.

Other seminal unsupervised contrastive learning frameworks include the Momentum Contrast (MoCo) family \cite{he2020momentum, chen2020improved, chen2021empirical}, Bootstrap Your Own Latent (BYOL) \cite{grill2020bootstrap}, and SimSiam \cite{chen2021exploring}. The MoCo family aims to address the practical challenge of requiring huge batch sizes for training contrastive learning frameworks. It achieves this by maintaining a dynamic dictionary of encoded representations from preceding mini-batches. This dictionary is updated with features from a slowly progressing, momentum-updated encoder, thereby providing a significant and consistent set of negative examples without the computational overhead of a massive batch. Conversely, BYOL and SimSiam attempt to alleviate the representation collapse issue by leveraging encoders with asymmetric architectures to prevent direct parameter copying.

While the aforementioned methods are designed for unlabeled data, Khosla et al. \cite{khosla2020supervised} proposed Supervised Contrastive Learning (SupCon) to leverage label information within contrastive learning. The objective then becomes to pull together the representations of all samples within a class cluster, while simultaneously pushing this cluster away from all other class clusters. SupCon was shown to consistently outperform standard cross-entropy loss, particularly in terms of representation quality.

The superior generalizability of contrastive learning frameworks has garnered increasing attention from researchers. Huang et al. \cite{huang2023towards} reveal three factors that contribute to that. They are the alignment of positive samples, divergence of class centers, and concentration of augmented data. The first two are related to learning objectives, and the data augmentation process determines the last one. Recently, Hieu and Ledent \cite{hieugeneralization} proposed a generalization analysis for supervised contrastive representation learning in the practical, non-i.i.d. setting, where a fixed pool of labeled data is recycled to form training tuples. They derive excess risk bounds using a novel theoretical framework inspired by U-statistics \cite{hoeffding1992class}.

Although contrastive learning has proven transformative for classification tasks, it is not directly applicable to tasks with continuous target variables. The main challenge lies in defining positive and negative pairs for regression tasks.

To this end, Li et al. \cite{li2022robust} introduced a group-aware contrastive network that minimizes intra-group variance and maximizes inter-group distances by selecting positive and negative sample pairs based on age groups. Similarly, Pitawela et al. \cite{pitawela2025cloc} use samples with the same label to construct positive samples and samples with different labels to construct negative samples. They also introduced a configurable hyperparameter to determine the distance among clusters in the feature space. Conversely, Zha et al. \cite{zha2024rank} define positive and negative pairs based on absolute age differences among samples. Specifically, for a given anchor sample, any other sample in the batch can form a positive pair, while the corresponding negative pairs are all samples whose label distance to the anchor is greater than that of the positive pair.

Compared to the aforementioned contrastive learning frameworks for regression tasks, OrdCon not only considers absolute age differences between samples but also leverages ordinal information between samples by contrasting direction vectors. This ensures that learned features are mapped in a direction that aligns with age progression, resulting in better generalizability.

\subsection{Order learning}

The idea of order learning was initially proposed by Lim et al \cite{lim2019order}. The authors introduced a new approach to problems with inherently ordered labels by focusing on relative comparisons rather than absolute value estimation. Instead of directly predicting a continuous value or performing multiple binary classifications, their method reframed the problem as determining the relative order between two instances. Later, Lee et al. \cite{lee2022geometric} integrated order learning with metric learning to create a feature space in which distances between instances reflect both the order and magnitude of age differences. Recently, Unsupervised Order Learning was proposed, which integrates the idea of order learning into clustering for ordinal data \cite{lee2024unsupervised}.

Different from the above order learning methods, OrdCon focuses on learning a generalized feature space by adopting contrastive learning to contrast direction vectors with age progression and features with age-specific proxies. Moreover, we tested its versatility on two age-related tasks rather than only one, like previous methods.

\subsection{Age Estimation}

Over the past few decades, various methods for face-based age estimation have been developed. Early approaches, like Kwon and Lobo's work \cite{kwon1994age}, classified faces into three age groups using craniofacial development theory and wrinkle analysis.

With the advent of larger age-labeled datasets \cite{ricanek2006morph, chen2014cross}, convolutional neural networks (CNNs) became the dominant feature extraction method. One of the early CNN-based approaches was \cite{wang2015deeply} in which a two-layer CNN was used for age prediction.

A primary challenge in age estimation is that age is not a discrete, independent class but an ordinal, continuous variable. To this end, Niu et al. \cite{niu2016ordinal} and Chen et al. \cite{Chen_2017_CVPR} framed age estimation as an ordinal regression problem, using a parallel set of fully connected layers to handle binary classification sub-problems for each age. Pan et al. \cite{pan2018mean} proposed a mean-variance loss to concentrate age predictions around the ground truth. Shen et al. \cite{shen2019deep} tackled inhomogeneity in the mapping between facial features and age by attaching deep forests to CNNs, allowing random forests to perform soft data partitioning and learn age distributions.

To mine the ordinal relationships among ages, label distribution learning (LDL) \cite{geng2013facial} as been widely adapted by the research community. In LDL, instead of using a single age label, it represents the ground truth as a soft probability distribution over a range of ages. This allows a single training sample to contribute to the learning of its chronological age, as well as the ages of its neighbors. Recent advances in LDL have focused on designing sophisticated loss functions. Specifically, Adaptive LDL \cite{geng2014facial, li2022unimodal} has been proposed to generate instance-aware, unimodal distributions reflecting sample-specific ambiguity. On the other hand, ordinal LDL \cite{wen2023ordinal} aims to explicitly models the sequential and semantic ordinality of age by incorporating spatial, semantic, and temporal order relationships into its loss function. Although LDL-based methods works well on standard benchmarks, they perform poorly on cross-dataset evaluations \cite{paplham2024call}.

Unlike the aforementioned age estimation methods, our approach primarily focuses on improving the generalizability of age estimation models by combining contrastive learning and order learning techniques. In addition, the proposed framework has been demonstrated to be able to tackle both age estimation and AIFR problems rather than just a single usage.

\subsection{Age-Invariant Face Recognition}\

The mainstream strategy in modern AIFR is feature disentanglement, which aims to separate identity-specific features from age-related characteristics. Zheng et al. \cite{zheng2017age} proposed a multi-task framework with one path for face recognition and another for age estimation, directly subtracting age features from global features to obtain age-invariant identity features. Wang et al. \cite{wang2018orthogonal} adopted a similar multi-task strategy, introducing a novel decomposition method to separate age features from identity features using a spherical coordinate system. They also applied a regression loss to learn finer age features, enhancing the decomposition process. Recently, Xie et al. \cite{xie2022implicit} proposed an Implicit and Explicit Feature Purification (IEFP) framework that uses a regularizer to minimize the mutual information between the identity-focused features and features from a pre-trained age-estimation network.

Another strategy gaining popularity is combining generative and discriminative models to form unified frameworks. The Age-Invariant Model (AIM) \cite{zhao2020towards} was one of the first to perform cross-age face synthesis and recognition simultaneously. Recently, Yao et al. \cite{yao2025synthetic} evaluated multiple age synthesis models and demonstrated that using synthesized faces at different ages can enhance the AIFR model's performance.

Like existing age estimation models, the above AIFR models are all designed for homogeneous-dataset experiments. Techniques to improve generalizability, such as contrastive learning frameworks, have been largely unexplored for AIFR. The only notable work is our preliminary work, CACon \cite{wang2024cross}, which maximized similarity between features from different age groups of the same identity using additional samples from a face synthesis model. However, this approach did not explicitly model the continuous progression of aging, which may lead to inaccurate feature representations due to entanglement between age and identity features.

In this work, we focus on generalized age feature extraction by initially learning age and identity features simultaneously. Once natural age progression is learned through the two contrastive objectives, the age-related variation is gradually minimized to obtain age-invariant identity features.

\begin{figure*}
\begin{center}
\includegraphics[width=1\textwidth]{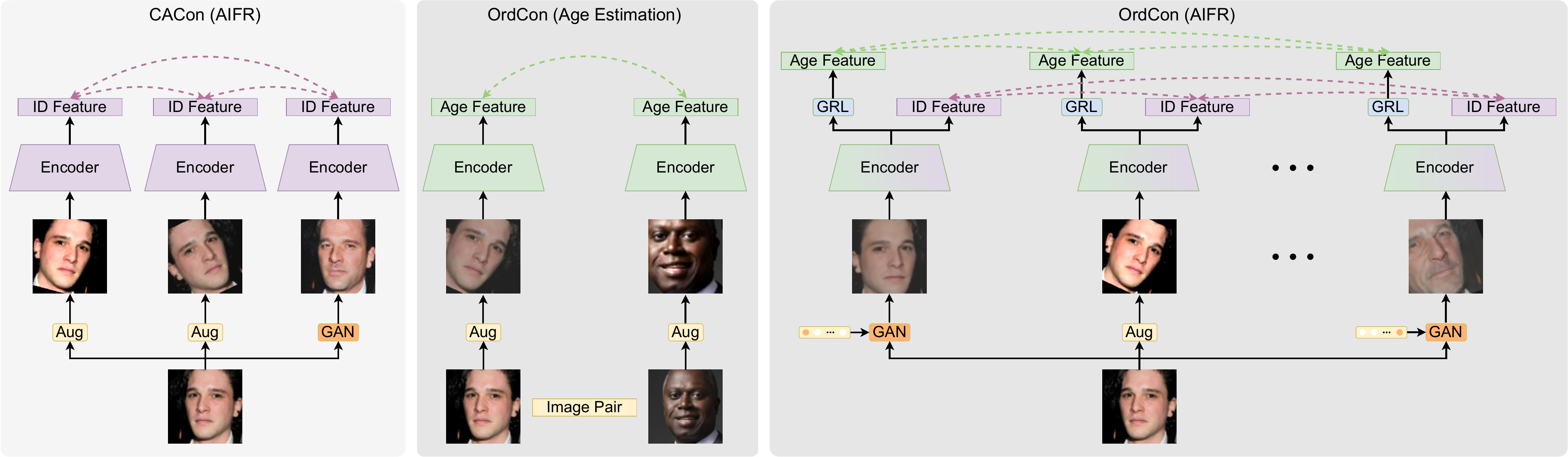}
\end{center}
   \caption{Proposed framework for age estimation and AIFR, compared to our preliminary work \cite{wang2024cross}. For age estimation, OrdCon processes an image pair with age differences to learn the ordinal relationship between images. For AIFR, CACon uses a third sample from a different age group to minimize the distances among samples from various age groups. Extending this approach, OrdCon leverages samples for each predefined age group to learn the individual's aging process in a multitask manner. To achieve age-invariant identity features, the gradient of age information is gradually reversed using a GRL.}
\label{fig:architecture}
\end{figure*}  

\section{Order-Enhanced Contrastive Learning}

In this section, we provide a detailed explanation of the proposed OrdCon, beginning with the formulation of the age feature extraction. We then explain the contrastive learning objectives, followed by a discussion on how the method can be applied to age estimation and AIFR. Fig. \ref{fig:architecture} shows the workflow of the proposed method.

\subsection{Problem Formulation}

To extract age features, let $\mathcal{X} = \{(x^{(i)}, y_{age}^{(i)})\}_{i=1}^N$ represent a dataset with $N$ samples, where $x^{(i)} \in \mathbb{R}^{H \times W \times C}$ is the $i$-th facial image of height $H$, width $W$, and $C$ color channels, and $y_{age}^{(i)} \in \mathbb{R}$ is the corresponding age label. The objective is to learn a mapping function $f_{age}: \mathcal{X} \to \mathcal{Z}_{age}$, where $\mathcal{Z}_{age} \in \mathbb{R}^{d_{age}}$ is a feature space of dimension $d_{age}$. The extracted age feature $z^{(i)}_{age} = f_{age}(x^{(i)})$ encodes information that reflects the natural progression of aging while ensuring generalization across unseen identities and scenarios.

After feature extraction, age estimation can be formulated as a regression problem, where the goal is to predict the chronological age $\hat{y}_{age}^{(i)} \in \mathbb{R}$ from a given facial image $x^{(i)}$. The model is trained to minimize the error between the predicted age $\hat{y}_{age}^{(i)}$ and the ground-truth age $y_{age}^{(i)}$, and the learned features should be positioned in the feature space align with the direction of the age progression.

In contrast, AIFR aims to preserve identity across different ages. Given a set of facial images $\{x^{(j)}\}_{j=1}^M$ of the same individual taken at different ages, the model should map these images to a shared identity feature space $\mathcal{Z}_{id}$, where identity remains invariant to age-related changes. Let $z^{(j)}_{id} = f_{id}(x^{(j)})$ represent the identity feature vector extracted from image $x^{(j)}$, where $f_{id}$ is the learned mapping function. To ensure identity preservation, the model must minimize the distance among the identity features $\{z^{(j)}_{id}\}_{j=1}^M$.

\subsection{Contrastive Learning Objectives}

\subsubsection{Feature Direction Contrast through Order Learning}

To ensure that the feature space $\mathcal{Z}_{age}$ captures the natural progression of aging, we first define age relationships between sample pairs and align the direction vectors of feature pairs with reference directions representing age progression or regression.

For two facial images, $x^{(i)}$ and $x^{(j)}$, where $y_{age}^{(i)} < y_{age}^{(j)}$, the age relationship is considered progressive, denoted by $x^{(i)} \prec x^{(j)}$. This means that $x^{(j)}$ represents an older face than $x^{(i)}$. The goal is to ensure that the age feature of the younger face, $z_{age}^{(i)}$, is positioned before the age feature of the older face, $z_{age}^{(j)}$, in the age feature space.

Conversely, if $y_{age}^{(i)} > y_{age}^{(j)}$, the age relationship is regressive, denoted by $x^{(i)} \succ x^{(j)}$, meaning $x^{(i)}$ represents an older face than $x^{(j)}$. In this case, the feature space should reflect the correct temporal ordering by placing $z_{age}^{(i)}$ ahead of $z_{age}^{(j)}$. Lastly, the order learning objective ignores pairs where $y_{age}^{(i)} = y_{age}^{(j)}$, as there is no order information to learn. Instead, the proxy-based contrastive objective minimizes the distance between age features for these images, ensuring that features of the same age are closely grouped.

Given that $z_{age}^{(i)}$ and $z_{age}^{(j)}$ are the extracted age features from $x^{(i)}$ and $x^{(j)}$, respectively, the direction vector between the two features is defined as:
\begin{equation}\label{direction}
\begin{aligned}
    v^{(i,j)}_{d} = \frac{z_{age}^{(i)}-z_{age}^{(j)}}{||z_{age}^{(i)}-z_{age}^{(j)}||}.
\end{aligned}
\end{equation}

Inspired by \cite{lee2022geometric}, we introduce forward and backward reference directions, modeled by proxies for each age or age group, to represent the progressive and regressive age relationships between facial images. Proxies represent the centers of each age or age group in the feature space and are denoted by $\mathcal{C} = \{c_a|a\in\mathcal{A}\}$, where $\mathcal{A}$ is the set of all distinct age labels in the dataset. If the age relationship between $x^{(i)}$ and $x^{(j)}$ is progressive, we have:
\begin{equation}\label{direction}
\begin{aligned}
    v_f^{(i,j)} = v(c_{y_{age}^{(i)}},c_{y_{age}^{(j)}}),
\end{aligned}
\end{equation}
\begin{equation}\label{direction}
\begin{aligned}
    v_b^{(i,j)} = v(c_{y_{age}^{(i)}},c_{y_{age}^{(i)}-1}),
\end{aligned}
\end{equation}
where $v_f^{(i,j)}$ represents the forward reference direction, and $v_b^{(i,j)}$ represents the backward reference direction. The order learning-based contrastive objective is to maximize the similarity between $v_f^{(i,j)}$ and $v^{(i,j)}_{d}$, aligning the direction vector with the direction of age progression, while minimizing the similarity between $v_b^{(i,j)}$ and $v^{(i,j)}_{d}$. This is achieved by optimizing the following loss function:
\begin{equation}\label{direction}
\begin{aligned}
   \mathcal{L}_{progressive} = -\sum_{\substack{i=1}}^N\sum_{\substack{j=1}}^N p(x^{(i)}\prec{x^{(j)}})\log q(x^{(i)}\prec{x^{(j)}}),
\end{aligned}
\end{equation}
where
\begin{equation}\label{direction}
\begin{aligned}
   p(x^{(i)}\prec{x^{(j)}}) = \frac{\mathbbm{1}_{(x^{(i)} \prec x^{(j)})}}{\sum_{\substack{k=1}}^N\mathbbm{1}_{(x^{(i)} \prec x^{(k)})}},
\end{aligned}
\end{equation}
and
\begin{equation}\label{direction}
\begin{aligned}
   q(x^{(i)}\prec{x^{(j)}}) = \frac{exp(v_f^{(i,j)T} \cdot v^{(i,j)}_{d}/{\tau})}{\sum_{\substack{k=1}}^{N}exp(v_b^{(i,k)T} \cdot v^{(i,k)}_{d}/{\tau})}.
\end{aligned}
\end{equation}

$\tau$ is a scaling hyperparameter that controls the sensitivity to differences between positive and negative pairs. $\mathbbm{1}_{(\cdot)}$ is an indicator that equals one if the condition in the parenthesis is true and zero otherwise.

In the case of a regressive age relationship between two facial images, the forward and backward reference directions are redefined as:
\begin{equation}\label{direction}
\begin{aligned}
    v_f^{(i,j)} = v(c_{y_{age}^{(i)}},c_{y_{age}^{(i)}+1}),
\end{aligned}
\end{equation}
\begin{equation}\label{direction}
\begin{aligned}
    v_b^{(i,j)} = v(c_{y_{age}^{(j)}},c_{y_{age}^{(i)}}).
\end{aligned}
\end{equation}

The order learning-based contrastive objective, in this case, aims to minimize the similarity between $v_f^{(i,j)}$ and $v^{(i,j)}_{d}$ while maximizing the similarity between $v_b^{(i,j)}$ and $v^{(i,j)}_{d}$ by optimizing the regressive counterpart:
\begin{equation}\label{direction}
\begin{aligned}
   \mathcal{L}_{regressive} = -\sum_{\substack{i=1}}^N\sum_{\substack{j=1}}^N p(x^{(i)}\succ{x^{(j)}})\log q(x^{(i)}\succ{x^{(j)}}),
\end{aligned}
\end{equation}
where
\begin{equation}\label{direction}
\begin{aligned}
   p(x^{(i)}\succ{x^{(j)}}) = \frac{\mathbbm{1}_{(x^{(i)} \succ x^{(j)})}}{\sum_{\substack{k=1}}^N\mathbbm{1}_{(x^{(i)} \succ x^{(k)})}},
\end{aligned}
\end{equation}
and
\begin{equation}\label{direction}
\begin{aligned}
   q(x^{(i)}\succ{x^{(j)}}) = \frac{exp(v_b^{(i,j)T} \cdot v^{(i,j)}_{d}/{\tau})}{\sum_{\substack{k=1}}^{N}exp(v_f^{(i,k)T} \cdot v^{(i,k)}_{d}/{\tau})}.
\end{aligned}
\end{equation}

The aggregated order learning-based contrastive objective is defined as the sum of both terms:
\begin{equation}\label{direction}
\begin{aligned}
   \mathcal{L}_{order} = \mathcal{L}_{progressive} + \mathcal{L}_{regressive}.
\end{aligned}
\end{equation}

It is important to note that no weighted hyperparameters are used, as both progressive and regressive relationships are equally important to learn.

By incorporating order learning, OrdCon learns not only that samples are different, but also how they differ in an ordinal sense (e.g., elder/younger). This imposes a meaningful structure on the learned feature space, aligning it with the natural progression of aging. Such a structured representation is inherently more generalizable than simply mapping features to labels without considering their underlying relationships.

\subsubsection{Soft Proxy Matching Loss}

While order learning-based objective ensures that the model captures the natural progression of aging by aligning age features with reference directions, it does not directly address intra-class variance or inter-class separation, which are keys to good generalizability \cite{huang2023towards}. To address this, we propose a new contrastive objective with proxies to cluster similar age features together while ensuring that features from different ages remain distinct.

As previously discussed, each distinct age $a \in \mathcal{A}$ is associated with a proxy $c_a \in \mathbb{R}^{d_{age}}$. In proxy-based representation learning, these proxies act as representative centers for each age in the feature space, pulling features of the same age closer to their proxy while pushing them away from proxies representing other ages.

Unlike the loss functions proposed in \cite{movshovitz2017no} and \cite{kim2020proxy}, which utilize Euclidean distances, we propose a proxy matching mechanism based on cosine similarities to better capture the direction of age features, which is crucial for modeling age progression. For each given $x^{(i)}$, the proxy matching loss function is formulated as:
\begin{equation}\label{direction}
\begin{aligned}
   \mathcal{L}_{pm}(x^{(i)}, \mathcal{C}) = \frac{exp(sim(x^{(i)}, c_{y_{age}^{(i)}})/{\tau})}{\sum_{\substack{z \in \mathcal{Z}}}exp(sim(x^{(i)}, c_{z})/{\tau})},
\end{aligned}
\end{equation}
where $sim(\cdot)$ represents the cosine similarity and $c_{y_{age}^{(i)}}$ indicates the corresponding proxy $x^{(i)}$ should approach to. $\mathcal{Z}$ is a set of proxies such that $c_{y_{age}^{(i)}} \cup \mathcal{Z} = \mathcal{A}$ and $c_{y_{age}^{(i)}} \cap \mathcal{Z} = \varnothing$.

All proxies are initialized randomly, learnable, and assigned using static proxy assignment \cite{movshovitz2017no}. Specifically, each sample is assigned to a corresponding proxy based on its age label. During training, both the model and the proxy locations are updated, while the proxy assignment remains fixed.

While cosine similarity effectively captures the direction of age features, it does not account for the magnitude of different age differences. From a generalization perspective, the distance between an age feature and a proxy with lower similarity should be increased more than that between those with higher similarity. To address this, we propose the soft proxy matching loss, which assigns a weight term to each negative pair in the denominator of the loss function in Eq. (13). The soft proxy matching loss is formulated as:
\begin{equation}\label{direction}
\begin{aligned}
   \mathcal{L}_{spm}(x^{(i)}, \mathcal{C}) = \frac{exp(sim(x^{(i)}, c_{y_{age}^{(i)}})/{\tau})}{\sum_{\substack{z \in \mathcal{Z}}}w(y_{age}^{(i)}, z)exp(sim(x^{(i)}, c_{z})/{\tau})},
\end{aligned}
\end{equation}
where $w(\cdot, \cdot)$ is a weight based on the absolute difference between the label and the age represented by the corresponding proxy. It is defined as:
\begin{equation}\label{direction}
\begin{aligned}
   w(y_{age}^{(i)}, z) = \frac{1}{1+exp(-|y_{age}^{(i)} - z|/{\max\limits_{\forall z' \in \mathcal{Z}} |y_{age}^{(i)} - z'|})}.
\end{aligned}
\end{equation}

The proxy-based objective is defined as the aggregation of soft proxy matching losses for all input samples:
\begin{equation}\label{direction}
\begin{aligned}
   \mathcal{L}_{proxy} = -\frac{1}{N}\sum_{\substack{i=1}}^N \mathcal{L}_{spm}(x^{(i)}, \mathcal{C}).
\end{aligned}
\end{equation}

This new loss function ensures features of the same age cluster are tightly clustered while pushing features of different ages apart proportionally to their age difference. This creates a more discriminative and well-organized feature space, which is crucial for generalization, especially in cross-dataset scenarios where data distributions might differ.

\begin{figure}[!t]
\centering
\includegraphics[width=3.4in]{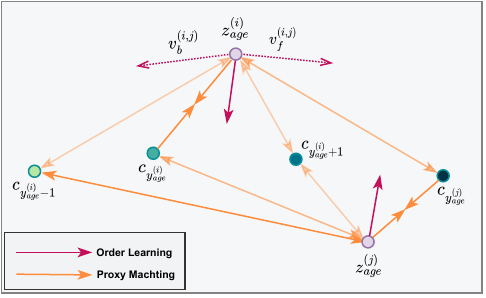}
\caption{Feature learning in OrdCon. The weights of push functions are represented by different levels of opacity, with higher opacity indicating a higher weight. Best viewed in color.}
\label{fig:feature}
\end{figure}

The overall contrastive loss is a weighted sum of the order learning objective and the proxy matching objective, formulated as:
\begin{equation}\label{direction}
\begin{aligned}
   \mathcal{L}_{contrast} = \mathcal{L}_{order} + \lambda\mathcal{L}_{proxy},
\end{aligned}
\end{equation}
where $\lambda$ is a weight factor that controls the relative importance of the two objectives. Fig. \ref{fig:feature} illustrates how age features representing different ages are pulled toward their corresponding proxies, with the learning of directions between features guided by reference directions.

\subsection{Application to Age Estimation and AIFR}

The proposed OrdCon is a new contrastive learning framework, and we apply it to two key age-related tasks: age estimation and AIFR. As shown in Fig.\ref{fig:architecture}, OrdCon learns age features that can be directly used for age estimation. Once age progression is modeled during the contrastive learning stage, a regression loss, such as L1 loss, can be used to map the learned features to their corresponding labels. For age estimation, the L1 loss is formulated as:
\begin{equation}\label{direction}
\begin{aligned}
   \mathcal{L}_{AE} = \frac{1}{N}\sum_{\substack{i=1}}^N |w_{fc}(z_{age}^{(i)}) - y_{age}^{(i)}|,
\end{aligned}
\end{equation}
where $w_{fc}$ represents the function learned by a fully connected layer.

For AIFR, on the other hand, since both identity and age features need to be learned, we use a multitask encoder with two fully-connected layers in parallel to learn these two sets of features simultaneously. Additionally, following \cite{wang2024cross}, age groups are used instead of specific ages, as usually there are no significant visual changes within a short time span.

Unlike \cite{wang2024cross}, where a generative adversarial network (GAN) \cite{wang2021age} synthesizes an additional sample within a random age group as part of the data augmentation process, OrdCon leverages samples for every age group except the one represented by the input face to learn the identity of the same individual across different ages. To achieve this, a condition (the age group label) is fed to the GAN to guide the synthesis process. We use the same GAN model as in \cite{wang2021age} due to its superior capability in identity preservation, which is crucial for face recognition with synthesized samples \cite{paraperas2024arc2face}.

To leverage multiple positive augmented samples from each input image, identity features are learned using the multi-positive contrastive loss \cite{tian2024stablerep}, formulated as:
\begin{equation}\label{direction}
\begin{aligned}
   \mathcal{L}_{id} = -\sum_{\substack{i=1}}^N\sum_{\substack{j=1}}^N p(y_{id}^{(i)}=y_{id}^{(j)})\log q(y_{id}^{(i)}=y_{id}^{(j)}),
\end{aligned}
\end{equation}
where
\begin{equation}\label{direction}
\begin{aligned}
   p(y_{id}^{(i)}=y_{id}^{(j)}) = \frac{\mathbbm{1}_{(y_{id}^{(i)}=y_{id}^{(j)})}}{\sum_{\substack{k=1}}^N\mathbbm{1}_{(y_{id}^{(i)}=y_{id}^{(k)})}},
\end{aligned}
\end{equation}
\begin{equation}\label{qid}
\begin{aligned}
   q(y_{id}^{(i)}=y_{id}^{(j)}) = \frac{exp(z_{id}^{(i)} \cdot z_{id}^{(j)}/{\tau})}{\sum_{\substack{k=1}}^{N}exp(z_{id}^{(i)} \cdot z_{id}^{(k)}/{\tau})},
\end{aligned}
\end{equation}
where $y_{id}^{(\cdot)}$ represents the identity label, and $z_{id}^{(\cdot)}$ denotes the extracted identity features.

The identity information learned in the conventional multi-task setting is entangled with age, which occludes age-invariant identity features. To address this, after a certain number of training epochs, the gradient of the age information is gradually reversed using a GRL, preventing the model from distinguishing between age groups and allowing it to learn age-invariant identity features. The gradient reversal during backpropagation can be formulated as:
\begin{equation}\label{qid}
\begin{aligned}
\frac{\partial \mathcal{L}_{contrast}}{\partial f_{grl}(x)} = -\lambda_{grl} \frac{\partial \mathcal{L}_{contrast}}{\partial x},
\end{aligned}
\end{equation}
where $f_{grl}(\cdot)$ represents the GRL function, and $\lambda_{grl}$ controls the strength of the gradient reversal.

\subsection{Generalizability Evaluation Matrices}

Based on the objectives of age estimation and AIFR mentioned in Section III.A, we adapt two matrices, alignment of positive samples and divergence of class centers, from \cite{huang2023towards} to test the generalizability of OrdCon. Alignment of positive samples measures the average squared Euclidean distance between feature vectors of two augmented views, and divergence of class centers is defined by the average dot product between centers of every pair of different classes. Since they are designed for classification problems with categorical data, they can not be used directly on tasks involving ordinal data. To this end, we evaluate the alignment of positive samples on AIFR and modify the divergence of class centers to make it applicable to age estimation. The alignment of positive samples for AIFR is formulated as:
\begin{equation}\label{align}
\begin{aligned}
 \mathcal{A}lign = \frac{1}{N}\sum_{\substack{i=1}}^N  \|f_{id}(A_1(x^{(i)})) - f_{id}(A_2(x^{(i)}))\|^2,
\end{aligned}
\end{equation}
where $A_1(x^{(i)})$ and $A_2(x^{(i)})$ are two augmented samples of $x^{(i)}$. A small alignment means positive pairs are closer in the feature space, which is desirable.

Regarding divergence, instead of calculating the dot product between class centers, we take the dot product between direction vectors and the overall age progression direction. The overall age progression direction in this case is defined by the direction vector of the youngest and eldest class centers. We call this new matrix the divergence of direction vectors, and it is formulated as:
\begin{equation}\label{div}
\begin{aligned}
 \mathcal{D}iv = \frac{1}{|\mathcal{P}|} \sum_{(i,j) \in \mathcal{P}} \left( (v^{(i,j)})^\top \cdot v_{overall} \right),
\end{aligned}
\end{equation}
where $\mathcal{P}$ denotes the set of all progressive pairs, $v_{overall}$ is the overall age progression direction and is defined as:
\begin{equation}\label{direction}
\begin{aligned}
    v_{overall} = \frac{c_{eldest} - c_{youngest}}{||c_{eldest} - c_{youngest}||}.
\end{aligned}
\end{equation}

$c_{eldest}$ and $c_{youngest}$ are the class centers of the eldest and youngest ages in the dataset. Instead of the smaller the better when measuring the divergence of class centers, a divergence of direction vectors close to 1 is ideal, which means the directions of progressive pairs are in the same direction as the overall aging trend in the feature space.

\section{Experiments}

\subsection{Datasets}

We evaluate OrdCon on three commonly used age-oriented benchmark datasets, which include age and identity labels suitable for both age estimation and AIFR tasks: the MORPH II dataset \cite{ricanek2006morph}, the FG-NET dataset \cite{cootes2008fg}, and the AgeDB dataset \cite{moschoglou2017agedb}. The statistics for these datasets are listed in Table I. For both tasks, we conduct both homogeneous-dataset and cross-dataset experiments to demonstrate the robustness and generalizability of our method. Some state-of-the-art methods use the IMDB-WIKI dataset \cite{rothe2018deep} for pre-training; however, we do not use it, as we observed no significant performance improvement with our proposed framework. When making comparisons, we explicitly indicate whether a method uses the IMDB-WIKI dataset for pre-training.

The MORPH II dataset contains over 55,000 facial images from approximately 13,000 subjects, with ages ranging from 16 to 77 and an average age of 33. The distribution of race labels is highly imbalanced, with more than 96\% of the subjects identified as either Black or White, while individuals from Asia and other regions make up the remainder. The male-to-female ratio is about 5.5:1. The images in the MORPH II dataset are mugshots, which differ from the in-the-wild images found in the other two datasets.

The FG-NET dataset consists of 1,002 facial images from 82 non-celebrity subjects, with each subject having more than 10 images taken over an extended period. The dataset includes variations in pose, illumination, and expression (PIE), making it suitable for evaluating the robustness of age estimation and AIFR models.

The AgeDB dataset contains 16,516 facial images of 570 celebrities, with ages ranging from 0 to 101 years. The images also exhibit variations in PIE. In our experiments, we use only the AgeDB dataset for celebrities, as other celebrity datasets have noisy labels and overlapping identities, making them unsuitable for cross-dataset evaluations. Notably, we utilize the AgeDB dataset only for cross-dataset evaluations and ablation studies, as its performance is less frequently reported compared to that of the MORPH II and FG-NET datasets.

\begin{table}
\ra{1.05}
\begin{center}
\label{datastats}
\caption{Statistics of Three Benchmark Datasets}
\begin{tabular}{l@{\hskip 0.35in}c@{\hskip 0.35in}c@{\hskip 0.35in}c}\toprule
\hfil{Dataset} & \hfil{\#images} & \hfil{\#subjects} & \hfil{age range}  \\ \midrule
\hfil{AgeDB} & \hfil{16,516} & \hfil{570} & \hfil{[0, 101]} \\
\hfil{FG-NET} & \hfil{1,002} & \hfil{82} & \hfil{[0, 69]} \\
\hfil{MORPH II} & \hfil{55,134} & \hfil{13,618} & \hfil{[16, 77]} \\
\bottomrule
\end{tabular}
\end{center}
\end{table}

\subsection{Experimental Settings}

\subsubsection{Data Pre-processing}

We use the open-source computer vision library dlib \cite{dlib09} for image preprocessing. Initially, 68 facial landmarks are detected in each image, which is then cropped based on eye positions and resized to \(128\times{128}\) pixels.

For general data augmentation in OrdCon, we apply random resizing and cropping, random horizontal flipping, and random changes in contrast, hue, brightness, and saturation.

\subsubsection{Data Partition}

For age estimation experiments with the MORPH II dataset, three commonly used experimental settings are applied. In the first setting, referred to as Setting I, following previous works \cite{wang2018fusion,pan2018mean, Chen_2017_CVPR, niu2016ordinal}, the dataset is randomly split into two subsets, with 80\% used for training and 20\% for testing, ensuring no identity overlap. We generate 10 different partitions (with the same ratio but different splits) and report the mean values. In the second setting (Setting II), to address the imbalance in race distribution, the dataset is divided into three subsets: S1, S2, and S3, maintaining a 1:1 ratio between Black and White subjects and a 3:1 ratio between Male and Female subjects \cite{li2019bridgenet, chen2019age, yi2014age, guo2011simultaneous}. The goal is to select all females while maximizing the overall size of the subset. We train the model on S1 and test it on S2+S3, then train it on S2 and test it on S1+S3, and report the average result. Lastly, in Setting III, to reduce the variance caused by race imbalance, 5,492 images of White individuals are selected \cite{rothe2018deep,wang2015deeply}. These images are randomly split into training (80\%) and testing (20\%) sets, and final results are obtained using 5-fold cross-validation to further minimize data distribution variance.

For age estimation experiments with the FG-NET dataset, we adopt the leave-one-person-out (LOPO) strategy \cite{xia2020multi, shen2019deep, geng2013facial, geng2006learning}. In each fold, the facial images of one subject are used for testing, while images of all other subjects are used for training. With 82 subjects, this results in 82 folds, and the final results are reported as the average across all folds.

For AIFR experiments with the FG-NET dataset, we evaluate using three different settings. First, we adopt the leave-one-image-out (LOIO) strategy, as in previous work \cite{hou2021disentangled}, where one image is used for testing while the remaining 1,001 images are used for fine-tuning. This process is repeated 1,002 times, and the average result is reported. Additionally, we follow the protocols from Megaface Challenge 1 (MF1) \cite{kemelmacher2016megaface} and Megaface Challenge 2 (MF2) \cite{nech2017level}.

For the MORPH II dataset, we use the partition strategy from \cite{zhao2019look, wang2018orthogonal, wang2019decorrelated}, where either 20,000 images from 10,000 subjects (Setting 1) or 6,000 images from 3,000 subjects (Setting 2) are used as the test set.

For the AgeDB dataset in both age estimation and AIFR experiments, we use the official data partition.

\subsubsection{Implementation Details}

For fair comparison with other methods, we use ResNet-50 \cite{he2016deep} as the feature extractor for both age estimation and AIFR experiments. Fully connected layers of size 2048 are used to generate age and identity features for contrastive objectives. During fine-tuning, an additional fully connected layer is added, with size 1 for age estimation or a size equal to the number of identities in the dataset for identity recognition. The batch size is set to 4,096 during pre-training and 512 during fine-tuning. For contrastive pre-training, we run 400 epochs for age estimation and 1,000 epochs for AIFR. After epoch 500 in AIFR, we gradually adjust $\lambda_{grl}$ using the following equation:
\begin{equation}\label{gammagrl}
\begin{aligned}
\lambda_{grl} = \frac{2}{1+exp(-\gamma t)} - 1,
\end{aligned}
\end{equation}
where $t$ represents the training epoch, and $\gamma$ controls the growth rate of $\lambda_{grl}$, and is empirically set to 10 for all AIFR experiments. $\lambda$ in Eq. (17) is empirically set to 0.8 for all experiments. Additionally, we use the LARS optimizer for multi-GPU training during pre-training and Stochastic Gradient Descent (SGD) for fine-tuning.

\begin{table}
\ra{1.20}
\begin{center}
\caption{Comparison of MAE values on the MORPH II dataset for age estimation under all three settings. Pretrain indicates whether the method uses the IMDB-WIKI dataset for pretraining. The best results are highlighted in bold, and the second-best results are underlined.}
\label{tab:morph1}
\begin{tabular}{p{0.128\textwidth}p{0.06\textwidth}p{0.06\textwidth}p{0.06\textwidth}p{0.065\textwidth}}\toprule
\hfil{Method} & \hfil{Pretrain} & \hfil{Setting I} & \hfil{Setting II} & \hfil{Setting III}\\
\midrule
\hfil{OR-CNN \cite{niu2016ordinal}} & \hfil{N} & \hfil{3.27} & \hfil{-} & \hfil{-} \\
\hfil{Ranking-CNN \cite{Chen_2017_CVPR}} & \hfil{N} & \hfil{2.96} & \hfil{-} & \hfil{-} \\
\hfil{DEX \cite{rothe2018deep}} & \hfil{N} & \hfil{3.25} & \hfil{-} & \hfil{-} \\
\hfil{Mean-Var Loss \cite{pan2018mean}} & \hfil{N} & \hfil{2.80} & \hfil{-} & \hfil{-} \\
\hfil{FusionNet \cite{wang2018fusion}} & \hfil{N} & \hfil{2.76} & \hfil{-} & \hfil{-} \\
\hfil{DRF \cite{shen2019deep}} & \hfil{N} & \hfil{-} & \hfil{3.47} & \hfil{2.80}\\
\hfil{ARAN \cite{chen2019age}} & \hfil{N} & \hfil{-} & \hfil{2.63} & \hfil{-} \\
\hfil{MSFCL-KL \cite{xia2020multi}} & \hfil{N} & \hfil{2.73} & \hfil{-} & \hfil{-} \\
\hfil{VDAL \cite{liu2020similarity}} & \hfil{N} & \hfil{2.57} & \hfil{-} & \hfil{-} \\
\hfil{ADPF \cite{wang2022improving}} & \hfil{N} & \hfil{2.54} & \hfil{2.56} & \hfil{2.71} \\
\hfil{Hier-Att \cite{hiba2023hierarchical}} & \hfil{N} & \hfil{2.53} & \hfil{-} & \hfil{-} \\
\hfil{GOL \cite{lee2022geometric}} & \hfil{N} & \hfil{ 2.51*} & \hfil{ 2.60*} & \hfil{ 2.17*}  \\
\hfil{RNC \cite{zha2024rank}} & \hfil{N} & \hfil{ 2.47*} & \hfil{ 2.45*} & \hfil{\underline{ 2.12*}} \\
\hfil{Mixture \cite{zhao2024mixture}} & \hfil{N} & \hfil{2.54} & \hfil{-} & \hfil{-} \\
\hfil{FaRL+MLP \cite{paplham2024call}} & \hfil{N} & \hfil{ 2.99*} & \hfil{ 2.80*} & \hfil{ 2.51*} \\
\hfil{LRA-GNN \cite{zhang2025lra}} & \hfil{N} & \hfil{-} & \hfil{-} & \hfil{2.21} \\
\hfil{MCGRL \cite{shou2025masked}} & \hfil{N} & \hfil{\textbf{ 2.10*}} & \hfil{\textbf{ 2.28*}} & \hfil{ 2.15*} \\
\midrule
\hfil{DEX \cite{rothe2018deep}} & \hfil{Y} & \hfil{2.68} & \hfil{-} & \hfil{-} \\
\hfil{Mean-Var Loss \cite{pan2018mean}} & \hfil{Y} & \hfil{2.79} & \hfil{-} & \hfil{-} \\
\hfil{DAG \cite{taheri2019use}} & \hfil{Y} & \hfil{2.87} & \hfil{-} & \hfil{-} \\
\hfil{BridgeNet \cite{li2019bridgenet}} & \hfil{Y} & \hfil{-} & \hfil{2.63} & \hfil{2.38} \\
\hfil{DCDL \cite{sun2021deep}} & \hfil{Y} & \hfil{2.62} & \hfil{2.45} & \hfil{-} \\
\midrule
\hfil{OrdCon (Ours)} & \hfil{N} & \hfil{\underline{2.21}} & \hfil{\underline{2.36}} & \hfil{\textbf{2.07}} \\
\bottomrule
\end{tabular}
\end{center}
\end{table}

\subsection{Age Estimation Results}

For age estimation experiments, we report the Mean Absolute Error (MAE), which measures the average absolute difference between the ground truth and the predicted age.

\subsubsection{Homogeneous-Dataset Results}

\begin{figure*}
\begin{center}
\includegraphics[width=1\textwidth]{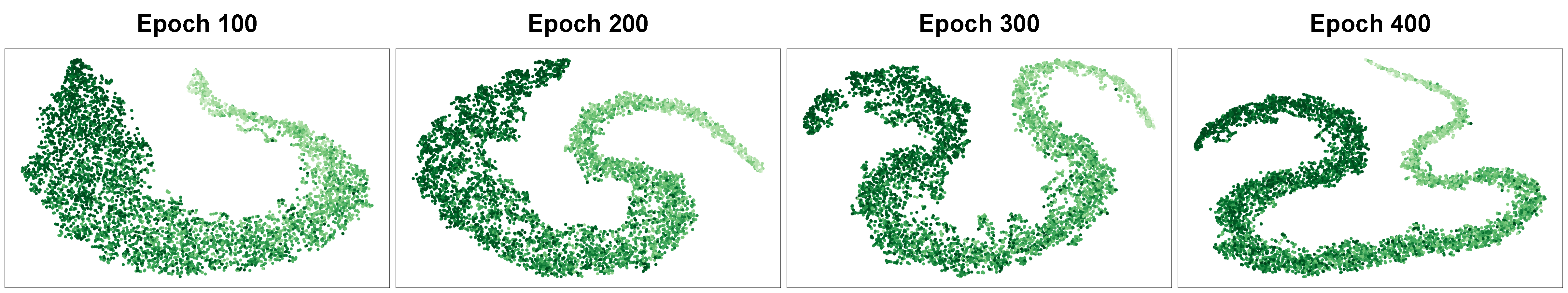}
\end{center}
   \caption{Learned age feature space at different training epochs on the MORPH II dataset. Darker colors indicate younger faces, while lighter colors indicate older faces. Best viewed in color.}
\label{fig:morph}
\end{figure*}

The homogeneous-dataset results for the MORPH II and FG-NET datasets are presented in Tables II and III, respectively. Note that in both tables, * indicates results produced by using the official code release. In Table II, results for all three settings are reported. As shown, OrdCon achieves either state-of-the-art or comparable results across all settings. Despite not using the IMDB-WIKI dataset for pretraining, our method outperforms those that do, highlighting the importance of modeling age progression between samples, whereas most methods only learn a direct mapping between features and labels. MCGRL \cite{shou2025masked}, which is also based on contrastive learning, achieves the lowest MAE in two of the three settings. One reason could be that it primarily focuses on learning the facial structure within the dataset, thereby gaining an advantage in homogeneous-dataset experiments. Notably, our method outperforms RNC \cite{zha2024rank}, which also employs contrastive learning to rank samples in order. By incorporating the soft proxy matching loss, our method learns features that are positioned closer to the center of each age, resulting in a more compact feature distribution that aligns with natural age progression. Fig. \ref{fig:morph} illustrates how the learned age feature space on the MORPH II dataset evolves during training. It also highlights the dataset's imbalance, with the majority of younger faces represented by the thicker dark green end and the minority of older faces represented by the lighter green end.

In Table III, the homogeneous-dataset results on the FG-NET dataset show a similar trend to those on the MORPH II dataset. However, due to the small size of the FG-NET dataset, methods pretrained with the IMDB-WIKI dataset achieve relatively lower MAE compared to those trained from scratch. For OrdCon, we found that it does not benefit from pretraining using IMDB-WIKI, as it could not effectively define progressive and regressive pairs based on the noisy labels in that dataset. Despite this, OrdCon achieves comparable results to those of other methods without requiring additional pretraining.

\begin{table}
\ra{1.20}
\begin{center}
\caption{Comparison of MAE values on the FG-NET dataset for age estimation. Pretrain indicates whether the method uses the IMDB-WIKI dataset for pretraining. The best results are highlighted in bold, and the second-best results are underlined.}
\label{tab:fg}
\begin{tabular}{p{0.128\textwidth}p{0.06\textwidth}p{0.04\textwidth}}\toprule
\hfil{Method} & \hfil{Pretrain} & \hfil{MAE}\\ \midrule
\hfil{DEX \cite{rothe2018deep}} & \hfil{N} & \hfil{4.63} \\
\hfil{Mean-Var Loss \cite{pan2018mean}} & \hfil{N} & \hfil{4.10} \\
\hfil{GA-DFL \cite{liu2017group}} & \hfil{N} & \hfil{3.93} \\
\hfil{ARAN \cite{chen2019age}} & \hfil{N} & \hfil{3.79} \\
\hfil{DRF \cite{shen2019deep}} & \hfil{N} & \hfil{3.47} \\
\hfil{ADPF \cite{wang2022improving}} & \hfil{N} & \hfil{2.86} \\
\hfil{RNC \cite{zha2024rank}} & \hfil{N} & \hfil{ 2.92*} \\
\hfil{FaRL+MLP \cite{paplham2024call}} & \hfil{N} & \hfil{ 3.15*} \\
\hfil{MCGRL \cite{shou2025masked}} & \hfil{N} & \hfil{ 2.76*} \\
\midrule
\hfil{DEX \cite{rothe2018deep}} & \hfil{Y} & \hfil{3.09} \\
\hfil{DAG \cite{taheri2019use}} & \hfil{Y} & \hfil{3.05} \\
\hfil{Mean-Var Loss \cite{pan2018mean}} & \hfil{Y} & \hfil{\underline{2.68}} \\
\hfil{BridgeNet \cite{li2019bridgenet}} & \hfil{Y} & \hfil{\textbf{2.56}} \\
\midrule
\hfil{OrdCon (Ours)} & \hfil{N} & \hfil{2.85} \\
\bottomrule
\end{tabular}
\end{center}
\end{table}

\begin{table*}
\begin{center}
\label{tab:cross}
\caption{Comparison of cross-dataset results, where MO represents the MORPH II dataset, FG represents the FG-NET dataset, and AG represents the AgeDB dataset. Reg indicates conventional regression-based methods, and Con indicates contrastive learning-based methods. The results from the methods used for comparison in this table are produced using officially released code. The best results are highlighted in bold, and the second-best results are underlined.}
\begin{tabular}{p{0.15\textwidth}p{0.05\textwidth}p{0.08\textwidth}p{0.08\textwidth}p{0.08\textwidth}p{0.08\textwidth}p{0.08\textwidth}p{0.08\textwidth}}
\toprule
\hfil{Method} & \hfil{Type} & \hfil{MO $\Rightarrow$ FG} & \hfil{MO $\Rightarrow$ AG} & \hfil{FG $\Rightarrow$ MO} & \hfil{FG $\Rightarrow$ AG} & \hfil{AG $\Rightarrow$ MO} & \hfil{AG $\Rightarrow$ FG} \\ \midrule
\hfil{OR-CNN \cite{niu2016ordinal}} & \hfil{Reg} & \hfil{17.47} & \hfil{12.82} & \hfil{14.35} & \hfil{19.83} & \hfil{7.05} & \hfil{17.40} \\
\hfil{DLDL-v2 \cite{gao2018age}} & \hfil{Reg} & \hfil{17.57} & \hfil{12.68} & \hfil{13.57} & \hfil{19.75} & \hfil{7.20} & \hfil{19.01} \\
\hfil{Mean-Var Loss \cite{pan2018mean}} & \hfil{Reg} & \hfil{15.81} & \hfil{12.93} & \hfil{15.44} & \hfil{19.21} & \hfil{7.33} & \hfil{17.43} \\
\midrule
\hfil{SimCLR \cite{chen2020simple}} & \hfil{Con} & \hfil{26.46} & \hfil{25.95} & \hfil{25.71} & \hfil{29.67} & \hfil{23.43} & \hfil{28.39} \\
\hfil{SupCon \cite{khosla2020supervised}} & \hfil{Con} & \hfil{17.58} & \hfil{13.76} & \hfil{16.83} & \hfil{22.14} & \hfil{7.36} & \hfil{19.30} \\
\hfil{RNC \cite{zha2024rank}} & \hfil{Con} & \hfil{\underline{12.30}} & \hfil{10.93} & \hfil{\underline{12.08}} & \hfil{\underline{13.82}} & \hfil{\underline{6.59}} & \hfil{\underline{14.59}} \\
\hfil{FaRL+MLP \cite{paplham2024call}} & \hfil{Con} & \hfil{13.75} & \hfil{\underline{9.82}} & \hfil{12.51} & \hfil{18.57} & \hfil{6.94} & \hfil{15.64} \\
\hfil{CLOC \cite{pitawela2025cloc}} & \hfil{Con} & \hfil{15.29} & \hfil{11.08} & \hfil{13.78} & \hfil{18.64} & \hfil{8.51} & \hfil{19.36} \\
\hfil{MCGRL \cite{shou2025masked}} & \hfil{Con} & \hfil{14.65} & \hfil{11.99} & \hfil{12.78} & \hfil{15.33} & \hfil{7.55} & \hfil{17.82} \\
\hfil{OrdCon (ours)} & \hfil{Con} & \hfil{\textbf{10.81}} & \hfil{\textbf{9.62}} & \hfil{\textbf{10.72}} & \hfil{\textbf{12.41}} & \hfil{\textbf{6.37}} & \hfil{\textbf{12.72}} \\
\bottomrule
\end{tabular}
\end{center}
\end{table*}

\begin{table}
\begin{center}
\label{tab:morph_homogenous}
\caption{Comparison of Rank-1 accuracy on the MORPH II dataset for AIFR under two settings. Cls represents conventional classification-based methods, and Con represents contrastive learning-based methods. The best results are highlighted in bold, and the second-best results are underlined.}
\begin{tabular}{p{0.128\textwidth}p{0.06\textwidth}p{0.06\textwidth}p{0.06\textwidth}}
\toprule
\hfil{Method} & \hfil{Type} & \hfil{Setting-1} & \hfil{Setting-2} \\ \midrule
\hfil{LF-CNN \cite{wen2016latent}} & \hfil{Cls} & \hfil{97.51} & \hfil{-} \\
\hfil{OE-CNN \cite{wang2018orthogonal}} & \hfil{Cls} & \hfil{98.55} & \hfil{98.67} \\
\hfil{DM \cite{shakeel2019deep}} & \hfil{Cls} & \hfil{98.67} & \hfil{-} \\
\hfil{AIM \cite{zhao2019look}} & \hfil{Cls} & \hfil{99.13} & \hfil{98.81} \\
\hfil{DAL \cite{wang2019decorrelated}} & \hfil{Cls} & \hfil{98.93} & \hfil{98.97} \\
\hfil{MT-MIM \cite{hou2021disentangled}} & \hfil{Cls} & \hfil{-} & \hfil{99.43} \\
\hfil{IEFP \cite{xie2022implicit}} & \hfil{Cls} & \hfil{\textbf{99.93}} & \hfil{\textbf{99.95}} \\
\midrule
\hfil{SimCLR \cite{chen2020simple}} & \hfil{Con} & \hfil{94.08} & \hfil{93.79} \\
\hfil{SupCon \cite{khosla2020supervised}} & \hfil{Con} & \hfil{96.44} & \hfil{94.68} \\
\hfil{CACon \cite{wang2024cross}} & \hfil{Con} & \hfil{99.57} & \hfil{99.52} \\
\hfil{OrdCon (Ours) } & \hfil{Con} & \hfil{\underline{99.83}} & \hfil{\underline{99.70}} \\
\bottomrule
\end{tabular}
\end{center}
\end{table}

\begin{table}
\begin{center}
\label{fg_homogenous}
\caption{Comparison of Rank-1 accuracy on the FG-NET dataset for AIFR under all three settings. Cls represents conventional classification-based methods, and Con represents contrastive learning-based methods. The best results are highlighted in bold, and the second-best results are underlined.}
\begin{tabular}{p{0.128\textwidth}p{0.06\textwidth}p{0.06\textwidth}p{0.06\textwidth}p{0.06\textwidth}}
\toprule
\hfil{Method} & \hfil{Type} & \hfil{LOIO} & \hfil{MF1} & \hfil{MF2}  \\ \midrule
\hfil{LF-CNN \cite{wen2016latent}} & \hfil{Cls} & \hfil{88.10} & \hfil{-} & \hfil{-} \\
\hfil{OE-CNN \cite{wang2018orthogonal}} & \hfil{Cls} & \hfil{-} & \hfil{58.21} & \hfil{53.26} \\
\hfil{DM \cite{shakeel2019deep}} & \hfil{Cls} & \hfil{92.23} & \hfil{-} & \hfil{-} \\
\hfil{AIM \cite{zhao2019look}} & \hfil{Cls} & \hfil{93.20} & \hfil{-} & \hfil{-} \\
\hfil{DAL \cite{wang2019decorrelated}} & \hfil{Cls} & \hfil{94.50} & \hfil{57.92} & \hfil{60.01} \\
\hfil{MT-MIM \cite{hou2021disentangled}} & \hfil{Cls} & \hfil{94.21} & \hfil{-} & \hfil{-} \\
\hfil{MTLFace \cite{huang2021age}} & \hfil{Cls} & \hfil{94.78} & \hfil{57.18} & \hfil{-} \\
\hfil{MFNR-LIAAD \cite{truong2023liaad}} & \hfil{Cls} & \hfil{95.11} & \hfil{60.11} & \hfil{-} \\
\hfil{IEFP \cite{xie2022implicit}} & \hfil{Cls} & \hfil{\underline{96.21}} & \hfil{-} & \hfil{-} \\
\hfil{ISF \cite{zhang2024age}} & \hfil{Cls} & \hfil{94.67} & \hfil{58.53} & \hfil{-} \\
\midrule
\hfil{SimCLR \cite{chen2020simple}} & \hfil{Con} & \hfil{90.36} & \hfil{54.00} & \hfil{52.52} \\
\hfil{SupCon \cite{khosla2020supervised}} & \hfil{Con} & \hfil{91.66} & \hfil{57.86} & \hfil{57.97} \\
\hfil{CACon \cite{wang2024cross}} & \hfil{Con} & \hfil{{94.61}} & \hfil{\underline{64.37}} & \hfil{\underline{64.94}} \\
\hfil{OrdCon (Ours) } & \hfil{Con} & \hfil{\textbf{96.88}} & \hfil{\textbf{68.30}} & \hfil{\textbf{65.36}} \\
\bottomrule
\end{tabular}
\end{center}
\end{table}

\begin{table*}
\begin{center}
\label{tab:cross}
\caption{Comparison of cross-dataset results, where MO indicates the MORPH II dataset, FG indicates the FG-NET dataset, and AG indicates the AgeDB dataset. Cls indicates conventional classification-based method and Con indicates contrastive learning-based method. Results from methods used for comparison in this table are produced by official released code.}
\begin{tabular}{p{0.15\textwidth}p{0.05\textwidth}p{0.08\textwidth}p{0.08\textwidth}p{0.08\textwidth}p{0.08\textwidth}p{0.08\textwidth}p{0.08\textwidth}}
\toprule
\hfil{Method} & \hfil{Type} & \hfil{MO $\Rightarrow$ FG} & \hfil{MO $\Rightarrow$ AG} & \hfil{FG $\Rightarrow$ MO} & \hfil{FG $\Rightarrow$ AG} & \hfil{AG $\Rightarrow$ MO} & \hfil{AG $\Rightarrow$ FG} \\ \midrule
\hfil{AIM \cite{zhao2019look}} & \hfil{Cls} & \hfil{77.68} & \hfil{75.03} & \hfil{53.39} & \hfil{49.20} & \hfil{76.44} & \hfil{71.53} \\
\hfil{MTLFace \cite{huang2021age}} & \hfil{Cls} & \hfil{79.64} & \hfil{75.15} & \hfil{58.18} & \hfil{49.60} & \hfil{75.36} & \hfil{70.75} \\
\hfil{IEFP \cite{xie2022implicit}} & \hfil{Cls} & \hfil{80.55} & \hfil{\underline{86.80}} & \hfil{61.44} & \hfil{54.17} & \hfil{\underline{85.99}} & \hfil{\underline{79.96}} \\
\midrule
\hfil{SimCLR \cite{chen2020simple}} & \hfil{Con} & \hfil{76.04} & \hfil{79.44} & \hfil{54.07} & \hfil{48.09} & \hfil{72.19} & \hfil{63.89} \\
\hfil{SupCon \cite{khosla2020supervised}} & \hfil{Con} & \hfil{79.88} & \hfil{81.85} & \hfil{61.08} & \hfil{48.83} & \hfil{74.35} & \hfil{68.50} \\
\hfil{CACon \cite{wang2024cross}} & \hfil{Con} & \hfil{\underline{82.03}} & \hfil{86.77} & \hfil{\underline{66.53}} & \hfil{\underline{58.12}} & \hfil{82.04} & \hfil{76.62} \\
\hfil{OrdCon (Ours)} & \hfil{Con} & \hfil{\textbf{82.91}} & \hfil{\textbf{87.52}} & \hfil{\textbf{70.09}} & \hfil{\textbf{60.97}} & \hfil{\textbf{87.78}} & \hfil{\textbf{81.40}} \\
\bottomrule
\end{tabular}
\end{center}
\end{table*}

\subsubsection{Cross-Dataset Results}

For cross-dataset experiments, we utilize all three datasets, MORPH II, FG-NET, and AgeDB, to form six training-testing pairs. Specifically, we train the model on one dataset and test it on one of the other two. The results are presented in Table IV. As shown, the comparison includes three conventional regression-based methods and seven contrastive learning-based methods, including OrdCon. For each dataset, we use the entire dataset for either training or testing. 

When using the MORPH II dataset as the test set, the MAE values are relatively lower, as the MORPH II dataset is considered relatively easy due to fewer PIE variations and a shorter time span per identity. However, due to the domain gap between MORPH II and the other two datasets, the MAE values in cross-dataset experiments are still slightly higher than those reported in the homogeneous-dataset evaluation.

Due to the small size of the FG-NET dataset, using it as the training set results in higher MAE values on the test set compared to using the other datasets for training.

Since SimCLR and SupCon are designed for classification tasks, their MAEs are significantly higher compared to regression-targeted contrastive learning methods, such as RNC and OrdCon. Both achieve lower MAE values compared to conventional regression-based methods, owing to the superior generalizability of contrastive learning \cite{huang2023towards}. It is worth noting that OrdCon outperforms MCGRL in all the scenarios, highlighting the importance of modeling the aging progression for improving the model's generalizability.

\begin{figure*}
\begin{center}
\includegraphics[width=1\textwidth]{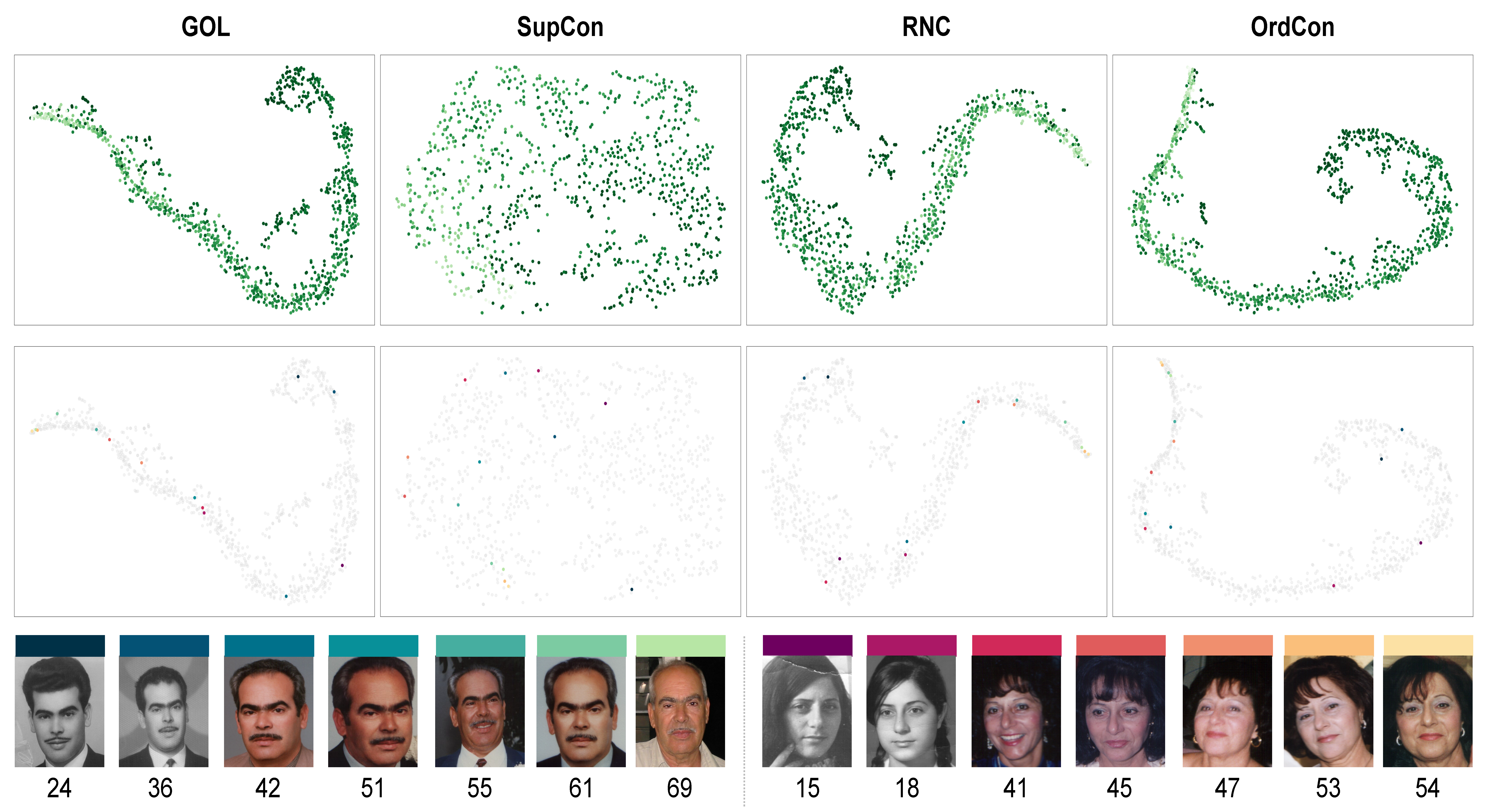}
\end{center}
   \caption{Comparison of learned age feature spaces from four methods, along with images of two sample identities. Best viewed in color and zoomed in.}
\label{fig:vis}
\end{figure*}

We compare the learned feature spaces of GOL, SupCon, RNC, and OrdCon in Fig.\ref{fig:vis}. The features are extracted from facial images in the FG-NET dataset using a model trained on the AgeDB dataset. Facial images from two sample identities are also shown at the bottom of the figure. Since SupCon is not designed for regression tasks, its learned feature space lacks a clear indication of age progression. In contrast, the feature spaces learned by GOL and RNC show a clear trend from younger to older faces. However, for some sample identities, the aging process does not always align with the general progression of aging.

GOL uses similar techniques to OrdCon, but without proxy-based representation learning and the soft proxy matching loss, features are pushed equally away from all other age clusters, which can lead to misplacement. While RNC considers age differences among features, it does not account for ordinal information, resulting in the absence of a clear aging direction. In comparison, the feature space learned by OrdCon shows a more distinct aging progression. 

\subsection{AIFR Results}

\subsubsection{Homogeneous-Dataset Results}

Similar to the homogeneous-dataset experiments on age estimation, we conduct experiments on the MORPH II and FG-NET datasets. In addition to our preliminary work \cite{wang2024cross}, CACon, we also include two commonly used contrastive learning methods, SimCLR and SupCon, for comparison to highlight the differences between our method and other contrastive approaches. The results on the two datasets are presented in Tables V and VI. As shown in Table V, OrdCon achieves the best results among contrastive learning-based methods in both settings and comparable results to state-of-the-art AIFR methods. However, it underperforms methods like IEFP \cite{xie2022implicit}, which leverages carefully designed purification units to learn dataset-specific correlation between age and identity features. Table VI shows that our method outperforms other AIFR methods on the FG-NET dataset across all three settings. Since OrdCon leverages one sample per age group for each facial image during training, it is less affected by the limited number of training samples in the FG-NET dataset. It also outperforms CACon, which uses only one additional sample at a random age group.

\subsubsection{Cross-Dataset Results}

Since there is limited work reporting results on cross-dataset AIFR experiments, for fair comparison, we only include methods with officially implementations, including three conventional classification-based methods and three contrastive learning-based methods. Again, in cross-dataset experiments, we use the entire dataset for either training or evaluation. The results are presented in Table VII. Due to the disentangling of identity features from robust and generalized age features in our method, OrdCon outperforms all conventional classification-based methods as well as contrastive learning-based methods that do not model natural age progression. Among conventional classification-based methods, IEFP achieves results most comparable to ours. Notably, IEFP and OrdCon share a similar strategy—learning age features first and then disentangling them from identity features.

\subsection{Generalizability Evaluation}
Apart from task-specific cross-dataset experiments, we also evaluate the generalizability of our framework against other contrastive learning frameworks using the alignment of positive samples and the divergence of direction vectors. For both matrices, we use the MOPRH II dataset as the training set, as it is the largest among the three we are using, and use the FG-NET dataset as the test set. As shown in Table VIII, OrdCon achieves the best results on both matrices. In terms of the alignment of positive samples, OrdCon outperforms CACon due to its aging progression modeling and the use of multiple augmented samples at different age groups during training. Furthermore, both achieve smaller alignment compared to general frameworks like SimCLR and SupCon. Regarding the divergence of direction vectors, OrdCon outperforms RNC because the latter only considers absolute age differences among samples and overlooks the ordinal relationships.

\begin{table}[t]
\ra{1.20}
\begin{center}
\caption{Evaluation of the generalizability for state-of-the-art methods and OrdCon.}
\label{tab:generalization}
\begin{tabular}{p{0.1\textwidth}p{0.06\textwidth}p{0.06\textwidth}}
\toprule
\hfil{Method} & \hfil{Align$\Downarrow$} & \hfil{Div$\Uparrow$}
\\ \midrule
\hfil{SimCLR \cite{chen2020simple}} & \hfil{0.77} & \hfil{-} \\
\hfil{SupCon \cite{khosla2020supervised}} & \hfil{0.63} & \hfil{-} \\
\hfil{CACon \cite{wang2024cross}} & \hfil{\underline{0.47}} & \hfil{-} \\
\hfil{RNC \cite{zha2024rank}} & \hfil{-} & \hfil{\underline{0.41}} \\
\hfil{CLOC \cite{pitawela2025cloc}} & \hfil{-} & \hfil{0.28} \\
\hfil{MCGRL \cite{shou2025masked}} & \hfil{-} & \hfil{0.32} \\
\midrule
\hfil{OrdCon (ours)} & \hfil{\textbf{0.40}} & \hfil{\textbf{0.56}} \\
\bottomrule
\end{tabular}
\end{center}
\end{table}

\begin{table}[t]
\ra{1.20}
\begin{center}
\caption{Evaluation of the contribution of each component in OrdCon on the AgeDB dataset for both age estimation and AIFR tasks.}
\label{tab:baseline}
\begin{tabular}{p{0.055\textwidth}p{0.04\textwidth}p{0.04\textwidth}p{0.045\textwidth}p{0.05\textwidth}p{0.05\textwidth}p{0.045\textwidth}}
\toprule
\hfil{contrast} & \hfil{order} & \hfil{proxy} & \hfil{MAE} & \hfil{Rank-1} & \hfil{Align$\Downarrow$} & \hfil{Div$\Uparrow$}
\\ \midrule
\hfil{\checkmark} & \hfil{} & \hfil{} & \hfil{9.32} & \hfil{91.03} & \hfil{0.63} & \hfil{-} \\
\hfil{\checkmark} & \hfil{\checkmark} &  &  \hfil{6.74} & \hfil{94.87} & \hfil{-} & \hfil{0.42} \\
\hfil{\checkmark} & & \hfil{hard} & \hfil{8.55} & \hfil{95.04} & \hfil{0.46} & \hfil{-} \\
\hfil{\checkmark} & \hfil{\checkmark} & \hfil{hard} & \hfil{6.15} & \hfil{97.21} & \hfil{0.42} & \hfil{0.53} \\
\hfil{\checkmark} & \hfil{\checkmark} & \hfil{soft} & \hfil{5.95} & \hfil{97.44} & \hfil{0.40} & \hfil{0.56} \\
\bottomrule
\end{tabular}
\end{center}
\end{table}

\begin{figure*}
\begin{center}
\includegraphics[width=1\textwidth]{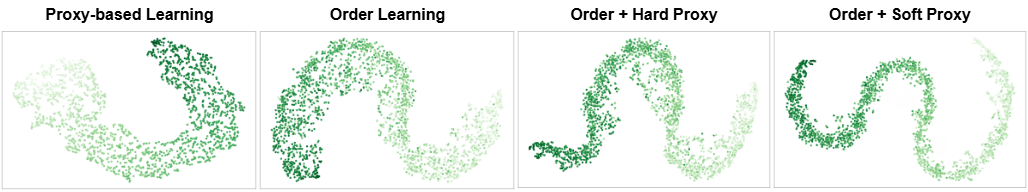}
\end{center}
   \caption{Comparison of age feature spaces on the AgeDB test set using different combinations of order learning and proxy-based learning.}
\label{fig:agedb}
\end{figure*}  

\subsection{Ablation Study}

\subsubsection{Contribution of Each Component}

We select SupCon \cite{khosla2020supervised} as the baseline as it has only a contrastive component and represents the fundamental form of supervised contrastive learning. In addition to this baseline model, we design three variations. The first two variations use either the order learning-based contrastive objective or the proxy-based contrastive objective independently to demonstrate their effectiveness compared to SupCon with a conventional contrastive learning objective. Next, we combine order learning with hard proxy-based learning using Eq. (13) to make the distribution of each age more compact, thereby improving performance. Finally, we replace hard proxy-based learning with its soft counterpart to create the complete OrdCon, making the proxy-based constraint proportional to age differences among samples for better modeling of age progression. 

Although all the frameworks in Table IX can be used for age estimation and AIFR to some extent, similar to Table VIII, we only evaluate alignment of positive samples on frameworks targeted on classification tasks, i.e., frameworks without the order learning component. Likewise, we only evaluate divergence of direction vectors on frameworks with the order learning component for regression tasks.

Fig. \ref{fig:agedb} compares the age feature spaces learned on the AgeDB test set using different combinations of order learning and proxy-based learning. The age progression becomes increasingly fine-grained as more objectives are added to the model. Notably, soft proxy matching loss alone can also roughly capture age progression. However, the features representing different ages are not as well separated as those learned with order learning.

\begin{table}
\ra{1.20}
\begin{center}
\caption{Comparison of Rank-1 accuracy under different age group granularities on the AgeDB dataset.}
\label{tab:heads}
\begin{tabular}{p{0.10\textwidth}p{0.035\textwidth}p{0.035\textwidth}p{0.035\textwidth}p{0.035\textwidth}p{0.035\textwidth}p{0.035\textwidth}}\toprule
\hfil{Granularity} & \hfil{5} & \hfil{6}& \hfil{7} & \hfil{8}& \hfil{9} & \hfil{10}\\ \midrule
\hfil{Rank-1} & \hfil{97.43} & \hfil{97.44} & \hfil{97.04} & \hfil{96.12} & \hfil{96.08} & \hfil{95.77}\\    
\bottomrule
\end{tabular}
\end{center}
\end{table}

\subsubsection{Age Group Granularity}

For AIFR, we also test how different age group partitions affect accuracy. Since facial aging effects are subtle over a few years and to control the computational complexity of our method, we test age group granularities ranging from 5 to 10 years. The results for different age group granularities on the AgeDB dataset are presented in Table X. The best results are achieved when the entire dataset is divided into groups with a 6-year age span (e.g., 0-5, 6-11, etc.). An age span of 5 years also yields similar results, albeit with higher computational complexity, as more samples need to be synthesized and utilized for contrastive learning. As the age span in each group increases, the AIFR accuracy gradually decreases due to higher intra-class dynamics.

\section{Discussion}

\subsection{Applicability}
While the current work focuses on age feature extraction for two mainstream age-related tasks, i.e., age estimation and AIFR, the underlying principles of OrdCon could be extended to other age-related tasks and facial attributes.

\subsubsection{To other age-related tasks}
Beyond age estimation and AIFR, OrdCon holds significant potential for other notable age-related tasks such as age synthesis and age privacy.

For age synthesis, the generalized age features extracted by OrdCon, which are specifically designed to capture the natural progression of aging, could guide fine-grained face aging synthesis. As highlighted by Yao et al. \cite{yao2025synthetic}, robust age representation is crucial for accurate age synthesis. By learning a feature space where age transitions are smooth and well-represented, OrdCon could guide generative models to produce more accurate facial images with rich aging features. More specifically, our learned age features, structured along an aging trajectory, could provide strong conditional input for GAN-based models or serve as a guide for latent space manipulation in diffusion models.

Regarding age privacy, the concept of extracting and potentially disentangling age features also has significant implications. Zhang et al. \cite{zhang2023rapp} discuss methods for concealing attributes while retaining utility. Similarly, suppose age features can be robustly identified and separated from other facial features. In that case, it might be possible to develop techniques to remove age information from facial images while preserving utility for other facial analysis tasks. Understanding how age is encoded in facial features is a prerequisite for effectively managing age-related privacy. A robust age feature extractor, such as OrdCon, is vital for these privacy-preserving transformations.

\subsubsection{To other attributes}
Based on Han et al. \cite{han2017heterogeneous}, facial attributes can be broadly categorized into ordinal and nominal attributes. Beyond age, OrdCon could be directly applicable to estimating other ordinal attributes, such as facial expression intensity \cite{zhao2016facial} or the more precise facial action unit intensity \cite{ma2024facial}. While nominal attributes like gender and race lack ordinal scales, they are often entangled with ordinal attributes such as age or hair length \cite{guo2009gender, guo2010study, albiero2021gendered}. In such cases, OrdCon could handle them similarly to how it addresses identity information in this work, i.e., by jointly learning both sets of features first and then reducing the variation in the nominal attribute caused by the ordinal attribute.

\subsection{Limitation}
\subsubsection{Generalizability Evaluation Matrix}
In this paper, we use two matrices, the alignment of positive samples and the divergence of direction vectors. The former is designed for categorical data and classification problems; hence, it does not apply to regression tasks with ordinal data. Also, we modified the divergence from measuring the averaged dot product of the class center to the direction vectors. The overall age progression direction in this work is defined as the direction from the youngest age to the eldest age, which may be inaccurate and not robust enough in cases where the distribution of age features does not follow a straight line. Therefore, more carefully designed evaluation matrices are needed for the analysis of the generalizability of contrastive learning frameworks on ordinal data.

\subsubsection{Homogeneous-dataset performance}
OrdCon's focus is primarily on extracting generalized features, which utilizes contrastive learning and order learning to model the aging progression in the feature space. That is the main reason it constantly outperforms other methods in cross-dataset experiments. However, its performance is surpassed by others' in some homogeneous-dataset experiments. Specifically, MCGRL outperforms OrdCon in some homogeneous-dataset experiments on age estimation, and IEFP outperforms our framework in some homogeneous-dataset experiments on AIFR. Both methods focus on feature extraction that is effective at extracting dataset-specific features. Specifically, MCGRL uses graphs to learn facial structures, which is hard to generalize if the demographics change from dataset to dataset. On the other hand, IEFP leverages carefully designed purification units to learn the correlation between age and identity features. This relationship is also hard to generalize as aging progress is highly personalized. Although they are not designed for generalizability, OrdCon could learn from these methods by combining its contrastive learning objectives with customized feature modeling techniques to thrive in both cross-dataset and homogeneous-dataset settings.

\section{Conclusion}

In this paper, we propose OrdCon to extract generalized age features for both age estimation and AIFR tasks. OrdCon is a new contrastive learning framework that leverages order learning to model natural age progression and a novel soft-proxy matching loss to minimize intra-class variance within each age cluster. The proposed method can be directly applied to age estimation. For AIFR, we employ a multitask feature extractor to simultaneously extract age and identity features, with a GRL used to make age information indistinguishable, thereby achieving age-invariant identity features. This paper also propose a new evaluation matrix to quantify the generalizability for regression methods. We demonstrate that our framework achieves either state-of-the-art or comparable performance in homogeneous-dataset experiments for both age estimation and the AIFR task. In addition, by utilizing contrastive learning objectives to improve generalizability, OrdCon outperforms other methods in cross-dataset experiments and generalizability evaluation matrices.

\section{Acknowledgment}
This work is supported by the EU Horizon 2020 – Marie Sklodowska-Curie Actions through the project Computer Vision Enabled Multimedia Forensics and People Identification (Project No. 690907, Acronym: IDENTITY).

\bibliographystyle{IEEEtran}
\bibliography{refs}

\begin{thebibliography}{10}
\providecommand{\url}[1]{#1}
\csname url@samestyle\endcsname
\providecommand{\newblock}{\relax}
\providecommand{\bibinfo}[2]{#2}
\providecommand{\BIBentrySTDinterwordspacing}{\spaceskip=0pt\relax}
\providecommand{\BIBentryALTinterwordstretchfactor}{4}
\providecommand{\BIBentryALTinterwordspacing}{\spaceskip=\fontdimen2\font plus
\BIBentryALTinterwordstretchfactor\fontdimen3\font minus \fontdimen4\font\relax}
\providecommand{\BIBforeignlanguage}[2]{{%
\expandafter\ifx\csname l@#1\endcsname\relax
\typeout{** WARNING: IEEEtran.bst: No hyphenation pattern has been}%
\typeout{** loaded for the language `#1'. Using the pattern for}%
\typeout{** the default language instead.}%
\else
\language=\csname l@#1\endcsname
\fi
#2}}
\providecommand{\BIBdecl}{\relax}
\BIBdecl

\bibitem{guo2009human}
G.~Guo, G.~Mu, Y.~Fu, and T.~S. Huang, ``Human age estimation using bio-inspired features,'' in \emph{2009 IEEE conference on computer vision and pattern recognition}.\hskip 1em plus 0.5em minus 0.4em\relax IEEE, 2009, pp. 112--119.

\bibitem{choi2011age}
S.~E. Choi, Y.~J. Lee, S.~J. Lee, K.~R. Park, and J.~Kim, ``Age estimation using a hierarchical classifier based on global and local facial features,'' \emph{Pattern recognition}, vol.~44, no.~6, pp. 1262--1281, 2011.

\bibitem{hu2016facial}
Z.~Hu, Y.~Wen, J.~Wang, M.~Wang, R.~Hong, and S.~Yan, ``Facial age estimation with age difference,'' \emph{IEEE Transactions on Image Processing}, vol.~26, no.~7, pp. 3087--3097, 2016.

\bibitem{xie2019chronological}
J.-C. Xie and C.-M. Pun, ``Chronological age estimation under the guidance of age-related facial attributes,'' \emph{IEEE Transactions on Information Forensics and Security}, vol.~14, no.~9, pp. 2500--2511, 2019.

\bibitem{fu2010age}
Y.~Fu, G.~Guo, and T.~S. Huang, ``Age synthesis and estimation via faces: A survey,'' \emph{IEEE transactions on pattern analysis and machine intelligence}, vol.~32, no.~11, pp. 1955--1976, 2010.

\bibitem{wang2020using}
H.~Wang, V.~Sanchez, W.~Ouyang, and C.-T. Li, ``Using age information as a soft biometric trait for face image analysis,'' \emph{Deep Biometrics}, pp. 1--20, 2020.

\bibitem{li2011discriminative}
Z.~Li, U.~Park, and A.~K. Jain, ``A discriminative model for age invariant face recognition,'' \emph{IEEE transactions on information forensics and security}, vol.~6, no.~3, pp. 1028--1037, 2011.

\bibitem{chen2015face}
B.-C. Chen, C.-S. Chen, and W.~H. Hsu, ``Face recognition and retrieval using cross-age reference coding with cross-age celebrity dataset,'' \emph{IEEE Transactions on Multimedia}, vol.~17, no.~6, pp. 804--815, 2015.

\bibitem{shakeel2019deep}
M.~S. Shakeel and K.-M. Lam, ``Deep-feature encoding-based discriminative model for age-invariant face recognition,'' \emph{Pattern Recognition}, vol.~93, pp. 442--457, 2019.

\bibitem{zhao2020towards}
J.~Zhao, S.~Yan, and J.~Feng, ``Towards age-invariant face recognition,'' \emph{IEEE Transactions on Pattern Analysis and Machine Intelligence}, vol.~44, no.~1, pp. 474--487, 2020.

\bibitem{park2010age}
U.~Park, Y.~Tong, and A.~K. Jain, ``Age-invariant face recognition,'' \emph{IEEE transactions on pattern analysis and machine intelligence}, vol.~32, no.~5, pp. 947--954, 2010.

\bibitem{sawant2019age}
M.~M. Sawant and K.~M. Bhurchandi, ``Age invariant face recognition: a survey on facial aging databases, techniques and effect of aging,'' \emph{Artificial Intelligence Review}, vol.~52, pp. 981--1008, 2019.

\bibitem{xia2020multi}
M.~Xia, X.~Zhang, L.~Weng, Y.~Xu \emph{et~al.}, ``Multi-stage feature constraints learning for age estimation,'' \emph{IEEE Transactions on Information Forensics and Security}, vol.~15, pp. 2417--2428, 2020.

\bibitem{rothe2018deep}
R.~Rothe, R.~Timofte, and L.~Van~Gool, ``Deep expectation of real and apparent age from a single image without facial landmarks,'' \emph{International Journal of Computer Vision}, vol. 126, no.~2, pp. 144--157, 2018.

\bibitem{wang2022improving}
H.~Wang, V.~Sanchez, and C.-T. Li, ``Improving face-based age estimation with attention-based dynamic patch fusion,'' \emph{IEEE Transactions on Image Processing}, vol.~31, pp. 1084--1096, 2022.

\bibitem{korban2023taa}
M.~Korban, P.~Youngs, and S.~T. Acton, ``Taa-gcn: A temporally aware adaptive graph convolutional network for age estimation,'' \emph{Pattern Recognition}, vol. 134, p. 109066, 2023.

\bibitem{akbari2020distribution}
A.~Akbari, M.~Awais, Z.-H. Feng, A.~Farooq, and J.~Kittler, ``Distribution cognisant loss for cross-database facial age estimation with sensitivity analysis,'' \emph{IEEE transactions on pattern analysis and machine intelligence}, vol.~44, no.~4, pp. 1869--1887, 2020.

\bibitem{akbari2021does}
A.~Akbari, M.~Awais, M.~Bashar, and J.~Kittler, ``How does loss function affect generalization performance of deep learning? application to human age estimation,'' in \emph{International Conference on Machine Learning}.\hskip 1em plus 0.5em minus 0.4em\relax PMLR, 2021, pp. 141--151.

\bibitem{huang2023towards}
W.~Huang, M.~Yi, X.~Zhao, and Z.~Jiang, ``Towards the generalization of contrastive self-supervised learning,'' in \emph{International Conference on Learning Representations}, 2023.

\bibitem{ramanathan2009age}
N.~Ramanathan, R.~Chellappa, S.~Biswas \emph{et~al.}, ``Age progression in human faces: A survey,'' \emph{Journal of Visual Languages and Computing}, vol.~15, pp. 3349--3361, 2009.

\bibitem{othmani2020age}
A.~Othmani, A.~R. Taleb, H.~Abdelkawy, and A.~Hadid, ``Age estimation from faces using deep learning: A comparative analysis,'' \emph{Computer Vision and Image Understanding}, vol. 196, p. 102961, 2020.

\bibitem{lim2019order}
K.~Lim, N.-H. Shin, Y.-Y. Lee, and C.-S. Kim, ``Order learning and its application to age estimation,'' in \emph{International Conference on Learning Representations}, 2019.

\bibitem{chen2020simple}
T.~Chen, S.~Kornblith, M.~Norouzi, and G.~Hinton, ``A simple framework for contrastive learning of visual representations,'' in \emph{International conference on machine learning}.\hskip 1em plus 0.5em minus 0.4em\relax PMLR, 2020, pp. 1597--1607.

\bibitem{khosla2020supervised}
P.~Khosla, P.~Teterwak, C.~Wang, A.~Sarna, Y.~Tian, P.~Isola, A.~Maschinot, C.~Liu, and D.~Krishnan, ``Supervised contrastive learning,'' \emph{Advances in neural information processing systems}, vol.~33, pp. 18\,661--18\,673, 2020.

\bibitem{zha2024rank}
K.~Zha, P.~Cao, J.~Son, Y.~Yang, and D.~Katabi, ``Rank-n-contrast: learning continuous representations for regression,'' \emph{Advances in Neural Information Processing Systems}, vol.~36, 2024.

\bibitem{wang2024cross}
H.~Wang, V.~Sanchez, and C.-T. Li, ``Cross-age contrastive learning for age-invariant face recognition,'' in \emph{ICASSP 2024-2024 IEEE International Conference on Acoustics, Speech and Signal Processing (ICASSP)}.\hskip 1em plus 0.5em minus 0.4em\relax IEEE, 2024, pp. 4600--4604.

\bibitem{ganin2015unsupervised}
Y.~Ganin and V.~Lempitsky, ``Unsupervised domain adaptation by backpropagation,'' in \emph{International conference on machine learning}.\hskip 1em plus 0.5em minus 0.4em\relax PMLR, 2015, pp. 1180--1189.

\bibitem{he2020momentum}
K.~He, H.~Fan, Y.~Wu, S.~Xie, and R.~Girshick, ``Momentum contrast for unsupervised visual representation learning,'' in \emph{Proceedings of the IEEE/CVF conference on computer vision and pattern recognition}, 2020, pp. 9729--9738.

\bibitem{chen2020improved}
X.~Chen, H.~Fan, R.~Girshick, and K.~He, ``Improved baselines with momentum contrastive learning,'' \emph{arXiv preprint arXiv:2003.04297}, 2020.

\bibitem{chen2021empirical}
X.~Chen, S.~Xie, and K.~He, ``An empirical study of training self-supervised vision transformers,'' in \emph{Proceedings of the IEEE/CVF international conference on computer vision}, 2021, pp. 9640--9649.

\bibitem{grill2020bootstrap}
J.-B. Grill, F.~Strub, F.~Altch{\'e}, C.~Tallec, P.~Richemond, E.~Buchatskaya, C.~Doersch, B.~Avila~Pires, Z.~Guo, M.~Gheshlaghi~Azar \emph{et~al.}, ``Bootstrap your own latent-a new approach to self-supervised learning,'' \emph{Advances in neural information processing systems}, vol.~33, pp. 21\,271--21\,284, 2020.

\bibitem{chen2021exploring}
X.~Chen and K.~He, ``Exploring simple siamese representation learning,'' in \emph{Proceedings of the IEEE/CVF conference on computer vision and pattern recognition}, 2021, pp. 15\,750--15\,758.

\bibitem{hieugeneralization}
N.~M. Hieu and A.~Ledent, ``Generalization analysis for supervised contrastive representation learning under non-iid settings,'' in \emph{International Conference on Machine Learning}, 2025.

\bibitem{hoeffding1992class}
W.~Hoeffding, ``A class of statistics with asymptotically normal distribution,'' \emph{Breakthroughs in statistics: Foundations and basic theory}, pp. 308--334, 1992.

\bibitem{li2022robust}
X.~Li, C.~Guo, Y.~Wu, C.~Zhu, and J.~Li, ``Robust age estimation model using group-aware contrastive learning,'' \emph{IET Image Processing}, vol.~16, no.~12, pp. 3201--3211, 2022.

\bibitem{pitawela2025cloc}
D.~Pitawela, G.~Carneiro, and H.-T. Chen, ``Cloc: Contrastive learning for ordinal classification with multi-margin n-pair loss,'' in \emph{Proceedings of the Computer Vision and Pattern Recognition Conference}, 2025, pp. 15\,538--15\,548.

\bibitem{lee2022geometric}
S.-H. Lee, N.~H. Shin, and C.-S. Kim, ``Geometric order learning for rank estimation,'' \emph{Advances in Neural Information Processing Systems}, vol.~35, pp. 27--39, 2022.

\bibitem{lee2024unsupervised}
S.-H. Lee, N.-H. Shin, and C.-S. Kim, ``Unsupervised order learning,'' in \emph{International Conference on Learning Representations}, 2024.

\bibitem{kwon1994age}
Y.~H. Kwon and N.~da~Vitoria~Lobo, ``Age classification from facial images,'' in \emph{Computer Vision and Pattern Recognition, IEEE Computer Society Conference on}, 1994, pp. 762--767.

\bibitem{ricanek2006morph}
K.~Ricanek and T.~Tesafaye, ``Morph: A longitudinal image database of normal adult age-progression,'' in \emph{Automatic Face \& Gesture Recognition, IEEE International Conference on}, 2006, pp. 341--345.

\bibitem{chen2014cross}
B.-C. Chen, C.-S. Chen, and W.~H. Hsu, ``Cross-age reference coding for age-invariant face recognition and retrieval,'' in \emph{European conference on computer vision}.\hskip 1em plus 0.5em minus 0.4em\relax Springer, 2014, pp. 768--783.

\bibitem{wang2015deeply}
X.~Wang, R.~Guo, and C.~Kambhamettu, ``Deeply-learned feature for age estimation,'' in \emph{2015 IEEE Winter Conference on Applications of Computer Vision}.\hskip 1em plus 0.5em minus 0.4em\relax IEEE, 2015, pp. 534--541.

\bibitem{niu2016ordinal}
Z.~Niu, M.~Zhou, L.~Wang, X.~Gao, and G.~Hua, ``Ordinal regression with multiple output cnn for age estimation,'' in \emph{Computer Vision and Pattern Recognition, IEEE Conference on}, 2016, pp. 4920--4928.

\bibitem{Chen_2017_CVPR}
S.~Chen, C.~Zhang, M.~Dong, J.~Le, and M.~Rao, ``Using ranking-cnn for age estimation,'' in \emph{Computer Vision and Pattern Recognition, IEEE Conference on}, July 2017.

\bibitem{pan2018mean}
H.~Pan, H.~Han, S.~Shan, and X.~Chen, ``Mean-variance loss for deep age estimation from a face,'' in \emph{Proceedings of the IEEE conference on computer vision and pattern recognition}, 2018, pp. 5285--5294.

\bibitem{shen2019deep}
W.~Shen, Y.~Guo, Y.~Wang, K.~Zhao, B.~Wang, and A.~L. Yuille, ``Deep differentiable random forests for age estimation,'' \emph{IEEE transactions on pattern analysis and machine intelligence}, 2019.

\bibitem{geng2013facial}
X.~Geng, C.~Yin, and Z.-H. Zhou, ``Facial age estimation by learning from label distributions,'' \emph{IEEE transactions on pattern analysis and machine intelligence}, vol.~35, no.~10, pp. 2401--2412, 2013.

\bibitem{geng2014facial}
X.~Geng, Q.~Wang, and Y.~Xia, ``Facial age estimation by adaptive label distribution learning,'' in \emph{2014 22nd International Conference on Pattern Recognition}.\hskip 1em plus 0.5em minus 0.4em\relax IEEE, 2014, pp. 4465--4470.

\bibitem{li2022unimodal}
Q.~Li, J.~Wang, Z.~Yao, Y.~Li, P.~Yang, J.~Yan, C.~Wang, and S.~Pu, ``Unimodal-concentrated loss: Fully adaptive label distribution learning for ordinal regression,'' in \emph{Proceedings of the IEEE/CVF Conference on Computer Vision and Pattern Recognition}, 2022, pp. 20\,513--20\,522.

\bibitem{wen2023ordinal}
C.~Wen, X.~Zhang, X.~Yao, and J.~Yang, ``Ordinal label distribution learning,'' in \emph{Proceedings of the IEEE/CVF International Conference on Computer Vision}, 2023, pp. 23\,481--23\,491.

\bibitem{paplham2024call}
J.~Paplh{\'a}m, V.~Franc \emph{et~al.}, ``A call to reflect on evaluation practices for age estimation: Comparative analysis of the state-of-the-art and a unified benchmark,'' in \emph{Proceedings of the IEEE/CVF Conference on Computer Vision and Pattern Recognition}, 2024, pp. 1196--1205.

\bibitem{zheng2017age}
T.~Zheng, W.~Deng, and J.~Hu, ``Age estimation guided convolutional neural network for age-invariant face recognition,'' in \emph{Proceedings of the IEEE Conference on Computer Vision and Pattern Recognition Workshops}, 2017, pp. 1--9.

\bibitem{wang2018orthogonal}
Y.~Wang, D.~Gong, Z.~Zhou, X.~Ji, H.~Wang, Z.~Li, W.~Liu, and T.~Zhang, ``Orthogonal deep features decomposition for age-invariant face recognition,'' in \emph{Proceedings of the European conference on computer vision (ECCV)}, 2018, pp. 738--753.

\bibitem{xie2022implicit}
J.-C. Xie, C.-M. Pun, and K.-M. Lam, ``Implicit and explicit feature purification for age-invariant facial representation learning,'' \emph{IEEE Transactions on Information Forensics and Security}, vol.~17, pp. 399--412, 2022.

\bibitem{yao2025synthetic}
W.~Yao, M.~A. Farooq, J.~Lemley, and P.~Corcoran, ``Synthetic face ageing: Evaluation, analysis and facilitation of age-robust facial recognition algorithms,'' \emph{IEEE Transactions on Biometrics, Behavior, and Identity Science}, 2025.

\bibitem{movshovitz2017no}
Y.~Movshovitz-Attias, A.~Toshev, T.~K. Leung, S.~Ioffe, and S.~Singh, ``No fuss distance metric learning using proxies,'' in \emph{Proceedings of the IEEE international conference on computer vision}, 2017, pp. 360--368.

\bibitem{kim2020proxy}
S.~Kim, D.~Kim, M.~Cho, and S.~Kwak, ``Proxy anchor loss for deep metric learning,'' in \emph{Proceedings of the IEEE/CVF conference on computer vision and pattern recognition}, 2020, pp. 3238--3247.

\bibitem{wang2021age}
H.~Wang, V.~Sanchez, and C.-T. Li, ``Age-oriented face synthesis with conditional discriminator pool and adversarial triplet loss,'' \emph{IEEE Transactions on Image Processing}, vol.~30, pp. 5413--5425, 2021.

\bibitem{paraperas2024arc2face}
F.~Paraperas~Papantoniou, A.~Lattas, S.~Moschoglou, J.~Deng, B.~Kainz, and S.~Zafeiriou, ``Arc2face: A foundation model for id-consistent human faces,'' in \emph{Proceedings of the European Conference on Computer Vision (ECCV)}, 2024.

\bibitem{tian2024stablerep}
Y.~Tian, L.~Fan, P.~Isola, H.~Chang, and D.~Krishnan, ``Stablerep: Synthetic images from text-to-image models make strong visual representation learners,'' \emph{Advances in Neural Information Processing Systems}, vol.~36, 2024.

\bibitem{cootes2008fg}
T.~Cootes and A.~Lanitis, ``The fg-net aging database,'' 2008.

\bibitem{moschoglou2017agedb}
S.~Moschoglou, A.~Papaioannou, C.~Sagonas, J.~Deng, I.~Kotsia, and S.~Zafeiriou, ``Agedb: the first manually collected, in-the-wild age database,'' in \emph{Proceedings of the IEEE Conference on Computer Vision and Pattern Recognition Workshop}, vol.~2, no.~3, 2017, p.~5.

\bibitem{dlib09}
D.~E. King, ``Dlib-ml: A machine learning toolkit,'' \emph{Journal of Machine Learning Research}, vol.~10, pp. 1755--1758, 2009.

\bibitem{wang2018fusion}
H.~Wang, X.~Wei, V.~Sanchez, and C.-T. Li, ``Fusion network for face-based age estimation,'' in \emph{2018 25th IEEE International Conference on Image Processing (ICIP)}.\hskip 1em plus 0.5em minus 0.4em\relax IEEE, 2018, pp. 2675--2679.

\bibitem{li2019bridgenet}
W.~Li, J.~Lu, J.~Feng, C.~Xu, J.~Zhou, and Q.~Tian, ``Bridgenet: A continuity-aware probabilistic network for age estimation,'' in \emph{Proceedings of the IEEE Conference on Computer Vision and Pattern Recognition}, 2019, pp. 1145--1154.

\bibitem{chen2019age}
Y.~Chen, S.~He, Z.~Tan, C.~Han, G.~Han, and J.~Qin, ``Age estimation via attribute-region association,'' \emph{Neurocomputing}, vol. 367, pp. 346--356, 2019.

\bibitem{yi2014age}
D.~Yi, Z.~Lei, and S.~Z. Li, ``Age estimation by multi-scale convolutional network,'' in \emph{Asian Conference on Computer Vision}.\hskip 1em plus 0.5em minus 0.4em\relax Springer, 2014, pp. 144--158.

\bibitem{guo2011simultaneous}
G.~Guo and G.~Mu, ``Simultaneous dimensionality reduction and human age estimation via kernel partial least squares regression,'' in \emph{CVPR 2011}.\hskip 1em plus 0.5em minus 0.4em\relax IEEE, 2011, pp. 657--664.

\bibitem{geng2006learning}
X.~Geng, Z.-H. Zhou, Y.~Zhang, G.~Li, and H.~Dai, ``Learning from facial aging patterns for automatic age estimation,'' in \emph{Proceedings of the 14th ACM international conference on Multimedia}, 2006, pp. 307--316.

\bibitem{hou2021disentangled}
X.~Hou, Y.~Li, and S.~Wang, ``Disentangled representation for age-invariant face recognition: A mutual information minimization perspective,'' in \emph{Proceedings of the IEEE/CVF International Conference on Computer Vision}, 2021, pp. 3692--3701.

\bibitem{kemelmacher2016megaface}
I.~Kemelmacher-Shlizerman, S.~M. Seitz, D.~Miller, and E.~Brossard, ``The megaface benchmark: 1 million faces for recognition at scale,'' in \emph{Proceedings of the IEEE conference on computer vision and pattern recognition}, 2016, pp. 4873--4882.

\bibitem{nech2017level}
A.~Nech and I.~Kemelmacher-Shlizerman, ``Level playing field for million scale face recognition,'' in \emph{Proceedings of the IEEE Conference on Computer Vision and Pattern Recognition}, 2017, pp. 7044--7053.

\bibitem{zhao2019look}
J.~Zhao, Y.~Cheng, Y.~Cheng, Y.~Yang, F.~Zhao, J.~Li, H.~Liu, S.~Yan, and J.~Feng, ``Look across elapse: Disentangled representation learning and photorealistic cross-age face synthesis for age-invariant face recognition,'' in \emph{Proceedings of the AAAI conference on artificial intelligence}, vol.~33, 2019, pp. 9251--9258.

\bibitem{wang2019decorrelated}
H.~Wang, D.~Gong, Z.~Li, and W.~Liu, ``Decorrelated adversarial learning for age-invariant face recognition,'' in \emph{Proceedings of the IEEE Conference on Computer Vision and Pattern Recognition}, 2019, pp. 3527--3536.

\bibitem{he2016deep}
K.~He, X.~Zhang, S.~Ren, and J.~Sun, ``Deep residual learning for image recognition,'' in \emph{Proceedings of the IEEE conference on computer vision and pattern recognition}, 2016, pp. 770--778.

\bibitem{liu2020similarity}
H.~Liu, P.~Sun, J.~Zhang, S.~Wu, Z.~Yu, and X.~Sun, ``Similarity-aware and variational deep adversarial learning for robust facial age estimation,'' \emph{IEEE Transactions on Multimedia}, vol.~22, no.~7, pp. 1808--1822, 2020.

\bibitem{hiba2023hierarchical}
S.~Hiba and Y.~Keller, ``Hierarchical attention-based age estimation and bias analysis,'' \emph{IEEE Transactions on Pattern Analysis and Machine Intelligence}, 2023.

\bibitem{zhao2024mixture}
Q.~Zhao, J.~Liu, and W.~Wei, ``Mixture of deep networks for facial age estimation,'' \emph{Information Sciences}, vol. 679, p. 121086, 2024.

\bibitem{zhang2025lra}
Y.~Zhang, Y.~Shou, W.~Ai, T.~Meng, and K.~Li, ``Lra-gnn: Latent relation-aware graph neural network with initial and dynamic residual for facial age estimation,'' \emph{Expert Systems with Applications}, p. 126819, 2025.

\bibitem{shou2025masked}
Y.~Shou, X.~Cao, H.~Liu, and D.~Meng, ``Masked contrastive graph representation learning for age estimation,'' \emph{Pattern Recognition}, vol. 158, p. 110974, 2025.

\bibitem{taheri2019use}
S.~Taheri and {\"O}.~Toygar, ``On the use of dag-cnn architecture for age estimation with multi-stage features fusion,'' \emph{Neurocomputing}, vol. 329, pp. 300--310, 2019.

\bibitem{sun2021deep}
H.~Sun, H.~Pan, H.~Han, and S.~Shan, ``Deep conditional distribution learning for age estimation,'' \emph{IEEE Transactions on Information Forensics and Security}, vol.~16, pp. 4679--4690, 2021.

\bibitem{liu2017group}
H.~Liu, J.~Lu, J.~Feng, and J.~Zhou, ``Group-aware deep feature learning for facial age estimation,'' \emph{Pattern Recognition}, vol.~66, pp. 82--94, 2017.

\bibitem{gao2018age}
B.-B. Gao, H.-Y. Zhou, J.~Wu, and X.~Geng, ``Age estimation using expectation of label distribution learning.'' in \emph{IJCAI}, vol.~1, 2018, p.~3.

\bibitem{wen2016latent}
Y.~Wen, Z.~Li, and Y.~Qiao, ``Latent factor guided convolutional neural networks for age-invariant face recognition,'' in \emph{Proceedings of the IEEE conference on computer vision and pattern recognition}, 2016, pp. 4893--4901.

\bibitem{huang2021age}
Z.~Huang, J.~Zhang, and H.~Shan, ``When age-invariant face recognition meets face age synthesis: A multi-task learning framework,'' in \emph{Proceedings of the IEEE/CVF conference on computer vision and pattern recognition}, 2021, pp. 7282--7291.

\bibitem{truong2023liaad}
T.-D. Truong, C.~N. Duong, K.~G. Quach, N.~Le, T.~D. Bui, and K.~Luu, ``Liaad: Lightweight attentive angular distillation for large-scale age-invariant face recognition,'' \emph{Neurocomputing}, vol. 543, p. 126198, 2023.

\bibitem{zhang2024age}
Z.~Zhang, S.~Yin, and L.~Cao, ``Age-invariant face recognition based on identity-age shared features,'' \emph{The Visual Computer}, vol.~40, no.~8, pp. 5465--5474, 2024.

\bibitem{zhang2023rapp}
Y.~Zhang, T.~Wang, R.~Zhao, W.~Wen, and Y.~Zhu, ``Rapp: Reversible privacy preservation for various face attributes,'' \emph{IEEE Transactions on Information Forensics and Security}, vol.~18, pp. 3074--3087, 2023.

\bibitem{han2017heterogeneous}
H.~Han, A.~K. Jain, F.~Wang, S.~Shan, and X.~Chen, ``Heterogeneous face attribute estimation: A deep multi-task learning approach,'' \emph{IEEE transactions on pattern analysis and machine intelligence}, vol.~40, no.~11, pp. 2597--2609, 2017.

\bibitem{zhao2016facial}
R.~Zhao, Q.~Gan, S.~Wang, and Q.~Ji, ``Facial expression intensity estimation using ordinal information,'' in \emph{Proceedings of the IEEE conference on computer vision and pattern recognition}, 2016, pp. 3466--3474.

\bibitem{ma2024facial}
B.~Ma, R.~An, W.~Zhang, Y.~Ding, Z.~Zhao, R.~Zhang, T.~Lv, C.~Fan, and Z.~Hu, ``Facial action unit detection and intensity estimation from self-supervised representation,'' \emph{IEEE Transactions on Affective Computing}, 2024.

\bibitem{guo2009gender}
G.~Guo, C.~R. Dyer, Y.~Fu, and T.~S. Huang, ``Is gender recognition affected by age?'' in \emph{2009 IEEE 12th International Conference on Computer Vision Workshops, ICCV Workshops}.\hskip 1em plus 0.5em minus 0.4em\relax IEEE, 2009, pp. 2032--2039.

\bibitem{guo2010study}
G.~Guo and G.~Mu, ``A study of large-scale ethnicity estimation with gender and age variations,'' in \emph{2010 IEEE Computer Society Conference on Computer Vision and Pattern Recognition-Workshops}.\hskip 1em plus 0.5em minus 0.4em\relax IEEE, 2010, pp. 79--86.

\bibitem{albiero2021gendered}
V.~Albiero, K.~Zhang, M.~C. King, and K.~W. Bowyer, ``Gendered differences in face recognition accuracy explained by hairstyles, makeup, and facial morphology,'' \emph{IEEE Transactions on Information Forensics and Security}, vol.~17, pp. 127--137, 2021.

\end{thebibliography}

\end{document}